\documentclass[lettersize,journal]{IEEEtran}
\usepackage{amsmath,amsfonts}
\usepackage{algpseudocode} 
\usepackage{algorithm}
\usepackage{algorithmicx}
\usepackage{algpseudocode} 
\usepackage{caption}
\usepackage{subfigure}
\usepackage{array}
\usepackage[caption=false,font=normalsize,labelfont=sf,textfont=sf]{subfig}
\usepackage{diagbox}
\usepackage{textcomp}
\usepackage{float}
\usepackage{picins}
\usepackage{multirow,booktabs}
\usepackage{verbatim}
\usepackage{graphicx}
\usepackage{cite}
\usepackage{soul} 
\usepackage{color, xcolor}

\usepackage[T1]{fontenc}
\usepackage{aecompl}

\begin{document}

\title{DK-SLAM: Monocular Visual SLAM with Deep Keypoint Learning, Tracking and Loop-Closing
}

\author{Hao Qu, Lilian Zhang, Jun Mao, Junbo Tie, Xiaofeng He, Xiaoping Hu, Yifei Shi, Changhao Chen*
\thanks{The authors are with the College of Intelligence Science and Technology, National University of Defense Technology, Changsha, 410073, China}
\thanks{*Corresponding author: Changhao Chen (changhao.chen66@outlook.com).}
\thanks{Hao Qu and Lilian Zhang contribute equally to this work.}
\thanks{This work was supported by National Natural Science Foundation of China (NFSC) under the Grant Number of 62103427, 62073331, 62103430, and Major Project of Natural Science Foundation of Hunan Province (No.2021JC0004). Changhao Chen is funded by the Young Elite Scientist Sponsorship Program by CAST (No. YESS20220181)}}

\markboth{In Submission,~Vol.~X, No.~X, June~2024}%
{Shell \MakeLowercase{\textit{et al.}}: A Sample Article Using IEEEtran.cls for IEEE Journals}


\maketitle

\begin{abstract}
The performance of visual SLAM in complex, real-world scenarios is often compromised by unreliable feature extraction and matching when using handcrafted features.  Although deep learning-based local features excel at capturing high-level information and perform well on matching benchmarks, they struggle with generalization in continuous motion scenes, adversely affecting loop detection accuracy. Our system employs a Model-Agnostic Meta-Learning (MAML) strategy to optimize the training of keypoint extraction networks, enhancing their adaptability to diverse environments. Additionally, we introduce a coarse-to-fine feature tracking mechanism for learned keypoints. It begins with a direct method to approximate the relative pose between consecutive frames, followed by a feature matching method for refined pose estimation.
To mitigate cumulative positioning errors, DK-SLAM incorporates a novel online learning module that utilizes binary features for loop closure detection. This module dynamically identifies loop nodes within a sequence, ensuring accurate and efficient localization. Experimental evaluations on publicly available datasets demonstrate that DK-SLAM outperforms leading traditional and learning based SLAM systems, such as ORB-SLAM3 and LIFT-SLAM. These results underscore the efficacy and robustness of our DK-SLAM in varied and challenging real-world environments.
\end{abstract}


\section{Introduction}
\IEEEPARstart{V}{isual} localization and mapping are fundamental components in autonomous systems for motion estimation and environmental perception. These capabilities have diverse applications, spanning from self-driving vehicles and unmanned aerial vehicles (UAVs) to mobile robots and immersive wearable technologies such as virtual reality (VR) and augmented reality (AR) devices. The quest for robust localization and mapping under varying environmental conditions has increasingly drawn the research community's attention.

Visual Simultaneous Localization and Mapping (SLAM) systems generally comprise two main components: front-end perception and back-end optimization. They are typically divided into two categories: direct methods and feature-based methods. Direct methods, such as LSD-SLAM \cite{engel2014lsd}, DSO \cite{DSO}, and SVO \cite{SVO}, process the pixel intensities in an image to optimize a photometric loss at the pixel level, directly using image data for map construction and pose estimation. While these methods can be highly accurate in specific scenarios, they often demand substantial computational resources and can become unstable in dynamic environments.

One the other hand, feature-based methods, exemplified by the ORB-SLAM series \cite{ORB-SLAM2, ORB-SLAM3}, focus on identifying and utilizing key image features, i.e. keypoints. These methods extract keypoints with distinctive properties, compute descriptors, and perform feature matching, outlier removal, and relative motion tracking. Their back-end processes include loop-closure detection and global optimization, crucial for refining and maintaining the system’s state estimates over time. Feature-based approaches typically excel in computational efficiency and robustness across diverse scenarios. However, they may struggle in environments with significant lighting changes or insufficient texture, where traditional handcrafted keypoints might not be detected or matched reliably.

In recent years, deep learning has demonstrated remarkable capabilities in computer vision tasks, such as image classification and object detection. Its integration into Visual Odometry (VO) and SLAM has garnered considerable interest. End-to-end methods use deep neural networks (DNNs) to directly infer pose estimates from images \cite{wang2017deepvo,DROID}. However, these methods often lack interpretability and may have difficulties generalizing across different scenarios due to discrepancies between training and testing data distributions. Learned-feature-based SLAM methods \cite{GCNv2, LIFT-SLAM, super-slam} replace traditional feature detection with deep learning-based visual feature extraction, which is then integrated into the SLAM back-end. These methods leverage both deep learning and geometric models to enhance SLAM performance.

Robust keypoint tracking and matching are vital for the reliability and effectiveness of visual SLAM systems in various applications. Although previous research has explored data-driven feature extraction in SLAM systems, there is still a need for more effective feature extraction networks tailored to the dynamic conditions encountered in SLAM applications. Current neural networks for feature extraction, such as SuperPoint \cite{superpoint}, are typically trained on static object detection datasets, which may limit their performance in dynamic, real-world environments with varying lighting and movement. Additionally, achieving reliable matching for learned features and efficient loop-closing remains a significant challenge. Unlike ORB features, which encapsulate explicit low-level image information, learned features often produce high-dimensional descriptor embeddings, complicating the process of establishing matches and filtering outliers. Moreover, enabling loop-closure detection using learned features poses further difficulties. Learning-based Bag-of-Words (BoW) methods for loop-closure detection involve complex training processes, often requiring extensive offline training and substantial memory storage, and they struggle with generalization to new environments.

To address the aforementioned challenges, we propose \textbf{DK-SLAM}, a novel \textbf{D}eep \textbf{K}eypoint based \textbf{SLAM} system, that incorporates deep keypoint extraction through meta-learning, coarse-to-fine feature tracking, and an efficient online learning-based loop-closing. Our DK-SLAM employs an enhanced Model-Agnostic Meta-Learning (MAML) strategy to extract robust deep local features, improving the system's ability to perform feature extraction in diverse and dynamic environments. This meta learning strategy enables our DK-SLAM system to adapt to diverse conditions without the need for additional training during deployment, thereby enhancing its applicability in real-world scenarios. Additionally, we introduce a novel feature matching strategy tailored for deep learned keypoints. This strategy involves a two-stage process: initial estimation of relative poses using patch photometric loss optimization, followed by refinement through 3D-2D relationships, resulting in improved matching accuracy and robustness. Furthermore, to enable more effective and efficient loop-closing, we propose an online learning-based Bag-of-Words (BoW) model. Our model compresses learned features into binary descriptors and utilizes images from previous timesteps within an online learning framework. It constructs a dynamic tree structure for bag-of-words, facilitating loop-closure detection in new environments. 

Experimental evaluation conducted on public car-driving and drone datasets, demonstrates the superior performance of our proposed DK-SLAM compared to representative traditional visual SLAM such as ORB-SLAM3 and LDSO, as well as learning-based systems such as LIFT-SLAM. Notably, compared to the representative monocular ORB-SLAM3, DK-SLAM achieves a translation accuracy improvement of approximately 17.7\% and a rotation accuracy enhancement of 24.2\% on the KITTI dataset, and it surpasses ORB-SLAM3 by approximately 34.2\% in translation accuracy on the EuRoC dataset.

Our contributions can be summarized as follows:

\begin{itemize}
   \item We propose DK-SLAM, a novel monocular visual SLAM system with deep keypoint meta learning. Our deep feature extractor, trained using a MAML strategy, enhances adaptability to diverse scenes.
   \item We develop a coarse-to-fine two-stage keypoint matching strategy that estimates relative poses through patch photometric loss optimization and refines them based on the 3D-2D relationship, resulting in improved accuracy.
 \item We introduce an online learning-based Bag-of-Words (BoW) model that effectively and efficiently utilizes binarized deep keypoints. This model dynamically fine-tunes itself to ensure accurate loop detection in the long-term operation.
\end{itemize}

The reminder of the paper is organized as follows: Section \ref{sec: related work} contains a survey of related work; Section \ref{sec: framework} introduces our proposed deep keypoint based monocular SLAM framework; Section \ref{sec: experiments} evaluates DK-SLAM in both car-driving and drone scenarios, and conducts extensive ablation studies; Section \ref{sec: conclusion} finally draws conclusions.

\section{Related Works}
In this section, we provide a brief overview of some related work in learning based feature extraction, learning based visual SLAM, and loop-closure detection.

\label{sec: related work}
\subsection{Learning based Feature Extractor}
Mainstream handcrafted extractors, including ORB \cite{ORB}, SIFT \cite{SIFT}, and Shi-Thomas \cite{ShiGoodFT}, are employed in visual SLAMs like ORB-SLAM2 \cite{ORB-SLAM2} and VINS-Mono \cite{Vins-Mono}. However, these methods relying on gradient information face challenges in dynamic lighting and low-texture environments.
To address these issues, researchers explored deep feature extractors. Superpoint \cite{superpoint} employs a shared encoder for high-level information and distinct decoders for detectors and descriptors, using synthesized labels for self-supervised training. D2-Net \cite{D2-Net} eliminates the need for a separate keypoint decoder, ensuring robustness. To enhance repeatability detection, \cite{Key.Net} combines multi-scale gradients with depth feature maps. Additionally, L2-Net \cite{L2-Net} introduces a local matching loss based on the L2 distance, improving feature matching performance. Experimental results demonstrate the effective feature matching performance of the learned local features on public datasets. Despite their success, the mentioned deep extractors exhibit inherent flaws in the face of diverse scenes, leading to potential performance degradation.

\begin{figure*}[t]
    \centering
    \includegraphics[width=18.2cm]{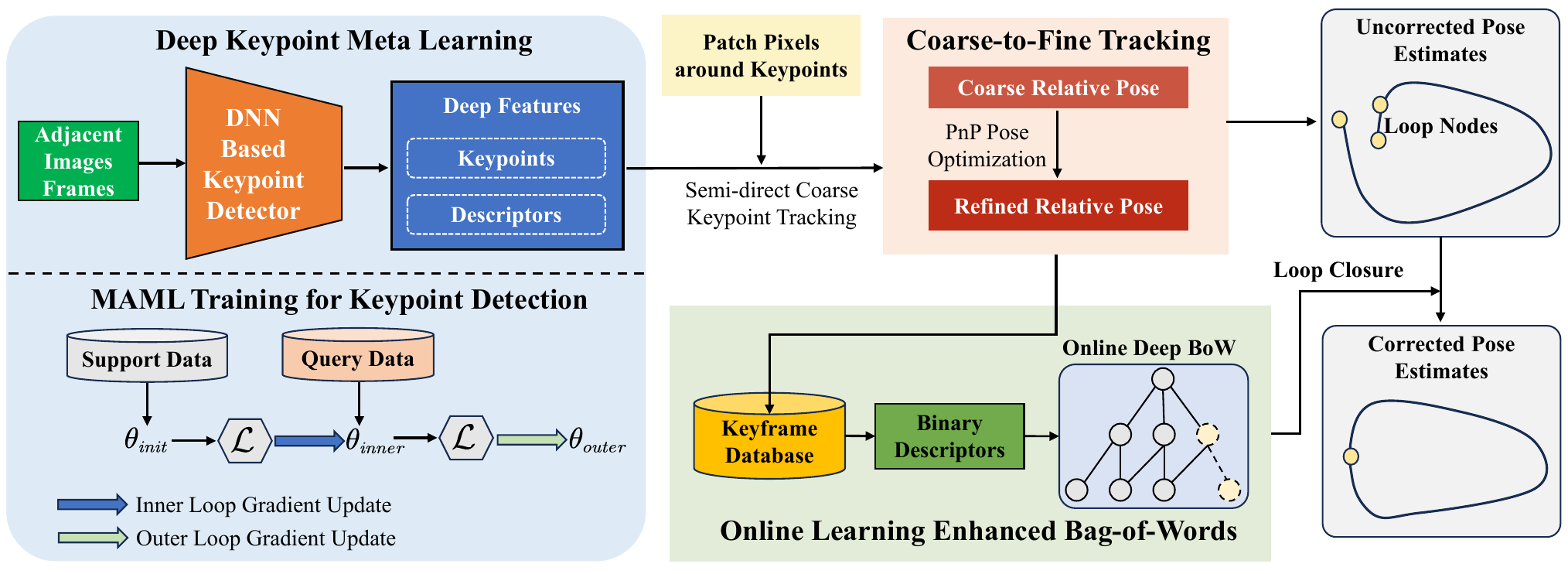}
    \caption{{An overview of our proposed DK-SLAM framework with deep keypoint meta learning, two-stage coarse-to-fine keypoint tracking and online learning based binary BoW for loop-closing.}}
    \label{fig: overview}
\end{figure*}

\subsection{Learning based Visual SLAM}
Recent research explores novel SLAM methods that exploit the advantages of both deep learning and multi-view geometry. GCNv2-SLAM integrates Graph Convolutional Networks (GCN) \cite{GCN} into SLAM, culminating in a comprehensive system with offline Bag-of-Words (BoW) training. In another study \cite{DXSLAM}, knowledge distillation from HF-Net \cite{From-croase} enhances feature detection within a compact model. Additionally, \cite{TIS} employs a self-supervised thermal optical flow tracking network, while \cite{SuperThermal} enhances Superpoint \cite{superpoint} with edge and multi-scale information. In \cite{LIFT-SLAM}, a LIFT \cite{LIFT} extractor serves as the SLAM front-end, implementing an adaptive feature matching strategy tailored to different datasets. In contrast, DK-SLAM adopts a meta-learning based training strategy, obviating the need for separate hyperparameter designs for distinct datasets. Moreover, \cite{TITS_SLAM} proposes a binary learning feature-based VSLAM that uses an adaptive motion model to provide more accurate initial pose estimates for tracking threads, achieving commendable localization performance on public datasets. To address the challenges posed by dynamic objects in visual SLAM, \cite{TIM_dynamic} introduces a semantic segmentation network that identifies dynamic objects in the scene. The system then removes feature points associated with these objects during the front-end processing, leading to state-of-the-art performance in environments with multiple moving objects. Additionally, \cite{TITS-object} exploits geometric information from objects to enhance VSLAM performance, and \cite{TITS-segmentation} applies learning-based semantic segmentation to correct scale in monocular SLAM mapping and localization.
Furthermore, \cite{TIM_Depth} incorporates learned depth labels into monocular visual SLAM, transforming it into a virtual stereo SLAM to overcome scale ambiguity. These depth labels are generated from unsupervised depth estimation networks, enabling the system to achieve more accurate mapping and localization.

\subsection{Loop Closure Detection}
Loop-closure detection is a crucial component of visual SLAM, tasked with identifying loop nodes and refining initial pose estimates \cite{TITS_loop}. Traditional Bag-of-Words (BoW) models like DBow2, DBow3, and FBow \cite{DBow2} utilize the binary descriptor BRIEF \cite{BRIEFBR} and store descriptors in a k-d tree structure based on Hamming distance. Despite their utility, these conventional BoW methods require extensive offline training data, which can limit their adaptability. To address this, iBow \cite{iBoW} was introduced, offering online BoW training directly from testing datasets. This approach provides robust transfer performance across various scenarios, enhancing the system's adaptability. Another approach to loop-closure detection, similar to NetVLAD \cite{NetVLAD}, as used in \cite{DXSLAM, super-slam}, depends heavily on the quality and relevance of the training set. Significant differences between the training and testing datasets can lead to performance degradation. In contrast,  \cite{TITS_loop2} utilizes deep learning to handle loop closure under varying viewpoints and lighting conditions. Instead of relying on the entire image, this approach focuses on landmarks to determine loop closure, enhancing its robustness and accuracy in diverse environments.


\section{Deep Keypoint based Monocular SLAM}
\label{sec: framework}
An overview of our proposed DK-SLAM system is illustrated in Figure \ref{fig: overview}. In the front-end, we utilize a neural network-based feature extractor to detect and describe keypoints across multi-scale image pyramids. To enhance generalization, the local feature extractor undergoes Model-Agnostic Meta-Learning (MAML) during training. To ensure a balanced distribution of keypoints across images of varying scales, we implement an averaging distribution strategy. Subsequently, to enhance the tracking and matching performance of the learned visual keypoints, we incorporate low-level information from the image. Initially, we construct local photometric constraints on the pixels surrounding keypoints to estimate the initial relative pose. Furthermore, we estimate the matching range of keypoints in adjacent frames based on this initial relative pose. Once the front-end completes keypoint tracking and pose estimation, keyframes are selected based on a predefined interval time scale. These selected keyframes are imported into the keyframe database, with the most recent keyframe being integrated into the back-end loop closure module. Memory consumption is optimized by converting floating keyframe descriptors into binary hash codes. Lastly, we introduce an online Bag-of-Words (BoW) module to establish a loop closure mechanism, facilitating closed-loop detection and global map optimization.

\subsection{Deep Keypoint Meta Learning}
\subsubsection{Feature Extractor Network}
Inspired by SuperPoint \cite{superpoint}, our deep keypoint extractor also leverages VGG16 \cite{VGG16} as the backbone architecture. Unlike SuperPoint, we introduce Batch Normalization layers after each Convolutional Neural Network (CNN) to encourage training convergence. After this backbone, there are decoders for keypoint detector  and descriptor, each consisting of multi-layer CNNs.
The first module, i.e. the detector decoder, produces a probability mask matching the input image size. We posit that keypoints are situated at positions where the mask value surpasses a certain threshold. The second module, i.e. the descriptor decoder, predicts feature descriptors for keypoints, which are then normalized to a dimension of 256 using the L2 norm.

\subsubsection{Self-Supervised Keypoint Learning}
Our network undergoes self-supervised training, employing a multi-step process. Initially, we obtain the homography-warped images from current training images. To train keypoint detector, we leverage the pre-trained MagicPoint network \cite{superpoint} to acquire keypoint pseudo labels for both the original and warped images. In both images, keypoint predictions and pseudo labels contribute to the formation of detector loss, denoted as ${L_p}$ and ${L_{wp}}$, with CrossEntropy serving as the loss metric. ${L_p}$ represents the detector loss of the original image, while $L_{wp}$ represents the detector loss of the warped image. Moving on to descriptor training, we utilize the sparse descriptor loss ${L_d}$ proposed in \cite{DeepKC}. This involves obtaining both the warped image and pixel-wise correspondences. Descriptor loss functions are then constructed solely for pseudo-labeled keypoints, employing triplet loss.
The sparse descriptor loss is composed of two components: homography-matched descriptor loss $L_{dm}$ and non-matched descriptor loss $L_{dn}$. The homography-matched descriptor loss, represented by Equation \ref{eq:loss_p}, involves a subscript $i$ ranging between $(0, 1, ..., N)$, where $(\mathbf{u}_i, \hat{\mathbf{u}_i})$ denotes the $N$ matched keypoint position pairs.

\begin{equation}
\label{eq:loss_p}
L_{dm}(\mathbf{u}_i) = {\left \| \mathbf{d}_I(\mathbf{u}_i) - \mathbf{d}_{wI}(\hat{\mathbf{u}_i} ) \right \| }^2
\end{equation}

As depicted in Equation \ref{eq:loss_d}, we use $M$ pairs of descriptors from the vicinity of matched descriptors to formulate a non-matched descriptor loss. The subscript $j$ ranges from 0 to $M$, where $(\mathbf{u}_i, \mathbf{u'}_j)$ represents unmatched keypoint position pairs.

\begin{equation}
\label{eq:loss_d}
L_{dn}(\mathbf{u}_i)=\frac{1}{M}\sum_{j=1}^{M}(\left \| \mathbf{d}_I(\mathbf{u}_i) - \mathbf{d}_{wI}(\mathbf{u'}_j)\right \|^2 )  
\end{equation}

The descriptor loss ${L_d}$, illustrated in Equation \ref{eq: Lrpw loss} with $z$ representing the boundary margin, is expressed as:

\begin{equation}
 \label{eq: Lrpw loss}
L_{d}=\frac{1}{N}\sum_{i=1}^{N}\text{max}(0, z+L_{dm}-L_{dn} ) 
\end{equation}

This descriptor loss ${L_d}$ aims to minimize the feature distance between matching descriptors while maximizing the feature distance between non-matching descriptors. The overall training loss is the sum of the aforementioned detector loss and descriptor loss. Here, $\lambda$ acts as a parameter to balance these different losses.

\begin{equation}
\label{eq: all loss fuction}
L_\text{all}= (L_p+L_{wp})+ \lambda{L_d} 
\end{equation}

\subsubsection{MAML-Based Visual Keypoint Meta Learning}
Learning-based visual keypoint (local feature) extractors face poor generalization due to variances in visual features across scenes. Significant feature disparity between training and unseen datasets can lead to catastrophic forgetting. To address this, we leverage insights from Model-Agnostic Meta-Learning (MAML), adapting network parameters through meta-training, which enhances generalization to new scenarios without requiring additional training during deployment. The MAML-based meta-training consists of inner loop training and outer loop training. The training set is partitioned into a support set $D_s$ and a query set $D_q$, where $D_s$ is involved in inner loop training, and $D_q$ participates in outer loop training.

In the MAML training strategy, the original network parameter $\mathbf{\theta}_a$ is initially duplicated to $\mathbf{\theta}_b$. Subsequently, both parameters undergo training in both the inner and outer loops. Within the inner loop, the support dataset $D_s$ is utilized for iterative updates to $\mathbf{\theta}_a$. This involves dividing a batch of support dataset into $n$ distinct tasks $(D^i_s, i=0...n-1)$, performing $m$ parameter updates on each task, resulting in the updated parameter $\mathbf{\theta}^m_a$. After completing one iteration of inner loop training, a minibatch of query set data $D^i_q$ is used to update the $\mathbf{\theta}_b$. The detailed steps of this feature meta-training strategy using MAML are shown in the Algorithm \ref{alg:Framwork} below.

\begin{algorithm}[htb]  
\caption{The Procedure of Deep Keypoint Meta Learning in DK-SLAM.}  
\label{alg:Framwork}  
\begin{algorithmic}[1]  
	\Require  
	Feature training dataset $D_f$, inner loop learning rate $\alpha_1$, outer loop learning rate $\alpha_2$. 
	\State Copy the initial network parameters as $\mathbf{\theta}_a$ and $\mathbf{\theta}_b$.  
	\label{code:fram:extract}  
	\While {not done}
        
         \State Sample a batch of support set $D_s$ and query set $D_q$ from $D_f$, whose batchsize are the same
        \State Sample a minibatch $D^i_s$ from $D_s$. Sample a minibatch $D^i_q$ from $D_q$. Where $i=0... n-1$. $n$ is the number of batchsize.
        \For{$D^i_s$ and $D^i_q$ in $D_s$, $D_q$}
        \State Calculate the gradient on $\mathbf{\theta}_a$ using $D^i_s$ and equation(4). The gradient is $\bigtriangledown_{\mathbf{\theta}_a}L_{all}$.
        \State Update $\mathbf{\theta}_a$ with $\bigtriangledown_{\mathbf{\theta}_a}L_{all}$.  $\mathbf{\theta}' _a=\mathbf{\theta}_a-\alpha _1\bigtriangledown_{\mathbf{\theta}_a}L_{all}$
        \State Repeat step 6 and 7 for m iterations. Received updated m times $\mathbf{\theta}^m_a$. In our implementation, we set m as 4. We call these iterations as inner loop. 
        \State Calculate the gradient on $\mathbf{\theta}^m_a$ using $D^i_q$ and equation(4). The gradient is $\bigtriangledown_{\mathbf{\theta}^m_a}L_{all}$.
        \State Update $\mathbf{\theta}_b$ with $\bigtriangledown_{\mathbf{\theta}^m_a}L_{all}$.  $\mathbf{\theta}' _b=\mathbf{\theta}_b-\alpha _2\bigtriangledown_{\mathbf{\theta}^m_a}L_{all}$ We call the updating for $\mathbf{\theta}_b$ as outer loop.
    \EndFor
    \EndWhile
\end{algorithmic}  
\end{algorithm}  

\subsubsection{The Distribution Strategy of Deep Keypoints}
To ensure a balanced dispersion of learned keypoints across all scales and corners within an image, we incorporate a keypoint distribution strategy during the feature extraction stage. We uniformly distribute the entire set of keypoints across a multi-scale image pyramid. Within each layer of the image pyramid, a predefined number of keypoints is distributed evenly within the grid of different regions on the image. Thus this strategy can prevent the concentration of learned keypoints in specific areas.

\begin{figure}[t]
    \centering
    \includegraphics[width=9.2cm]{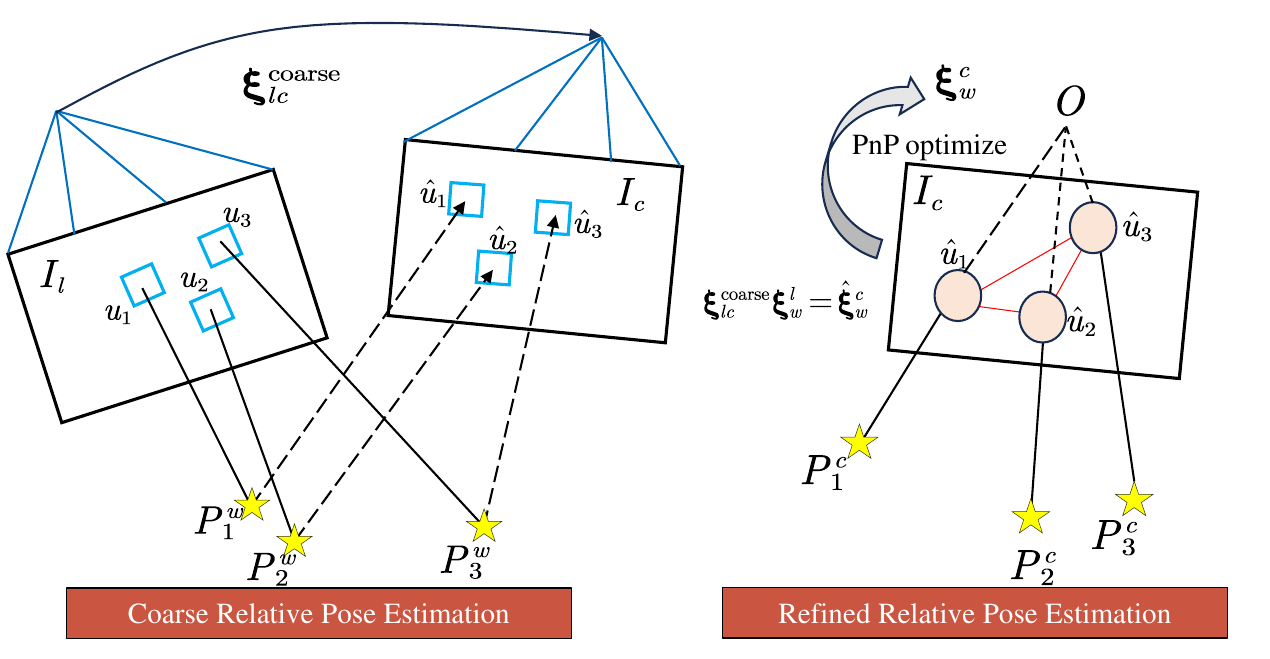}
    \caption{Diagram of our proposed coarse-to-fine two-stage keypoint tracking strategy. This process begins with relative pose estimation through patch photometric loss optimization, followed by refinement using the 3D-2D keypoint relationship for enhanced accuracy.}
    \label{overview}
\end{figure}

\subsection{Coarse-to-Fine Keypoint Tracking}
Accurate keypoint matching is essential for visual SLAM performance. Traditional systems like ORB-SLAM assume uniform motion between adjacent frames, often leading to challenges in finding correct matches and complicating optimization when this assumption fails. Unlike ORB features, which are straightforward low-level image descriptors, learned features generate complex high-dimensional embeddings, making matching and outlier filtering more difficult. To address these issues, we propose a two-stage deep keypoint tracking strategy to enhance matching robustness and accuracy.

\subsubsection{Semi-direct Coarse Keypoint Tracking}
Taking inspiration from the semi-direct visual odometry method \cite{SVO}, we introduce a coarse tracking method based on photometric constraints. We assume that the light intensity and texture around matching keypoints are similar in adjacent image frames. Map points from the last frame can be projected onto the current frame using the relative pose. If the relative pose is accurate, the photometric loss of the projected keypoint's patch will be minimal. As depicted in Equations \ref{eq: Photo loss1} and \ref{eq: Photo loss2}, the pixels of the keypoint's patch, the image points in the last frame coordinate $\mathbf{p}^{l}_i$, and the relative pose of adjacent frames $\mathbf{\xi}^\text{coarse}_{lc}$ collectively form the photometric loss $L_p$. Given the known map points and patch pixel values of the last frame, we adjust the relative pose $\mathbf{\xi}^\text{coarse}_{lc}$ to minimize the photometric loss $L_p$.

\begin{equation}
\label{eq: Photo loss1}
\mathbf{\xi}^\text{coarse}_{lc}=\arg \min _{\mathbf{\xi}^{\text{coarse}}_{lc}} \frac{1}{2} \sum_{i \in \overline{\mathcal{\chi}}}\left\|{L_p}\left({\mathbf{\xi}}^{\text{coarse}}_{lc}, \mathbf{p}^{l}_{i}\right)\right\|^{2}
\end{equation}

\begin{equation}
\label{eq: Photo loss2}
{L_p}\left(\mathbf{\xi}^{\text{coarse}}_{lc}, \mathbf{p}^{l}_{i}\right)=\mathbf{I}_{c}\left(\pi\left({\mathbf{\xi}}^{\text{coarse}}_{lc} \cdot \mathbf{p}^{l}_{i}\right)\right)-\mathbf{I}_{l}\left(\pi\left(\mathbf{p}^{l}_{i}\right)\right)
\end{equation}

Here, $\pi$ denotes the projection of the 3D map point from camera coordinates to pixel coordinates. $\mathbf{\overline{\mathcal{\chi}}}$ represents the set of keypoint indices. The photometric loss is constructed using the patches around the projected points.

\subsubsection{Coarse-to-Fine Keypoint Tracking}
Having obtained the coarse relative pose in the initial stage, we proceed to match map points between adjacent frames and refine the pose of the current frame $\mathbf{\xi}^{c}_{w}$ through the 3D-2D projection relationship. Subsequently, leveraging the coarse relative pose $\mathbf{\xi}^{\text{coarse}}_{lc}$, we project the 3D map points from the last frame's coordinate system to the pixel coordinate system of the current frame. Subsequently, we search for matching keypoints within a fixed radius range centered on the projection location, utilizing the Hamming distance of the descriptors as the search criterion.

Upon establishing the 3D-2D matching relationships, we construct a pose graph. As expressed in Equation \ref{eq: initial current pose}, we employ the coarse relative pose in conjunction with the last frame's pose to derive the initial pose of the current frame. 

\begin{equation}
\label{eq: initial current pose}
\hat{\mathbf{\xi}}^{c}_{w}= \mathbf{\xi}^{\text{coarse}}_{lc} \mathbf{\xi}^{l}_{w}
\end{equation}

The current frame's pose serves as the vertex in the graph, with the number of vertices being subject to optimization. Map points from the current frame and the reprojection loss of these map points constitute the edges. We exclusively optimize the pose within the tracking module. As detailed in Equation \ref{eq: reprojection loss}, we iteratively refine the current frame pose $\hat{\mathbf{\xi}}^{c}_{w}$ to minimize the reprojection loss. In our implementation, the g2o \cite{G2O} framework is employed for optimizing this pose graph.

\begin{equation}
\label{eq: reprojection loss}
{\mathbf{\xi}}^{c}_{w}=\arg \min _{\hat{\mathbf{\xi}}^{c}_{w}} \frac{1}{2} \sum_{i \in \mathbf{\overline{\mathcal{\chi}}}}\left\|{u_i}-\pi(\mathbf{\hat{\mathcal{\xi}}}^{c}_{w}.\mathbf{p}^{w}_i)\right\|^{2}
\end{equation}

\begin{figure}[t]
    \centering
    \includegraphics[width=9.2cm]{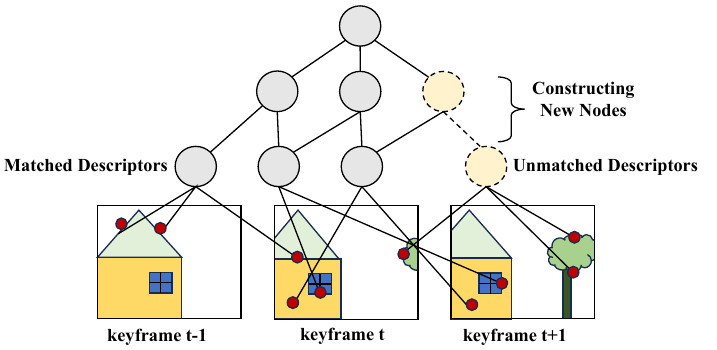}
    \caption{The illustration of our proposed Online Learning based Binary BoW. The BoW is constructed incrementally, with matched descriptors in the keyframes database stored within the same leaf node. In the presence of unmatched descriptors in the current keyframe, a new leaf node is created.}
    \label{online_tree}
\end{figure}

\subsection{Deep Keypoint Based Loop Closing}
Our DK-SLAM system features a deep-keypoint-based loop closure module that performs both loop-closure detection and correction. To improve generalization in unfamiliar environments, we utilize an online learning-based Bag-of-Words (BoW) model for effective loop-closure detection. Once loop closure is detected, the system identifies the closed-loop nodes and optimizes the global map using the relative poses from the detected loops.

\subsubsection{Online Learning for Binary BoW}
Unlike handcrafted descriptors, learned descriptors occupy a larger feature space, e.g. 128 dimensions in our case. Offline-trained BoW models may struggle to capture this space, making it difficult to distinguish deep feature indices in BoW leaf nodes. To tackle this, we introduce an online learning-based BoW model that uses extracted features exclusively from testing scene data. Constructing a tree-like structure with floating descriptors is challenging due to the large memory and time consumption, so, inspired by \cite{drones}, we employ binary hash transformation for processing deep features. As depicted in Equation \ref{eq: Online Deep Bow}, for descriptor vector values $\mathbf{d}_i$ less than 0, the processed $\mathbf{\hat{d}}_i$ is set to 0. For descriptor values greater than 0, the processed $\mathbf{\hat{d}}_i$ is set to 1, where $i$ ranges from 0 to 256.

\begin{figure*}[h]
	\centering
 \subfigure[Sequence 00]{ 
	\centering
	\includegraphics[width=0.315\linewidth]{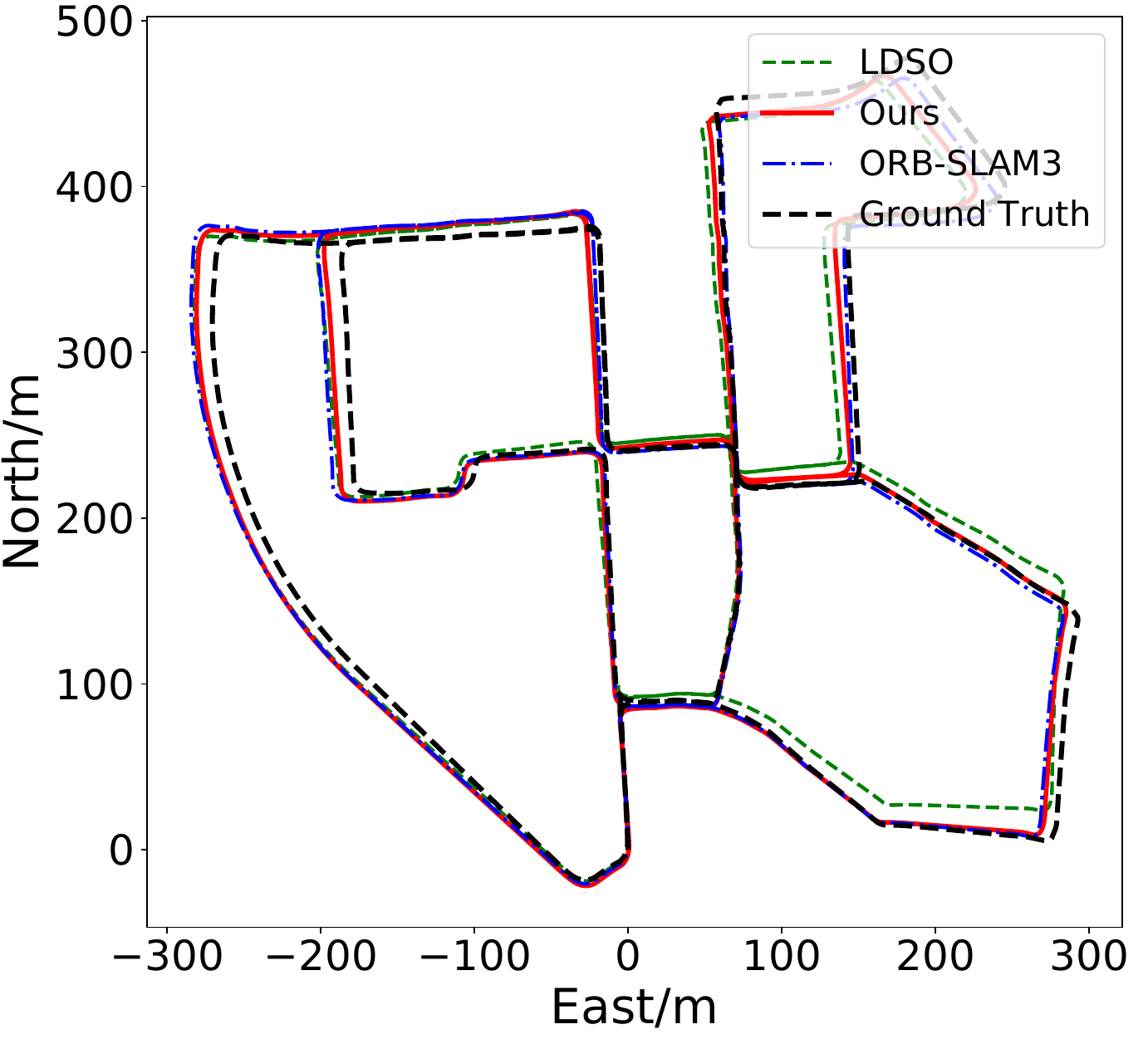}
 }
  \subfigure[Sequence 02]{ 
    \centering
    \includegraphics[width=0.315\linewidth]{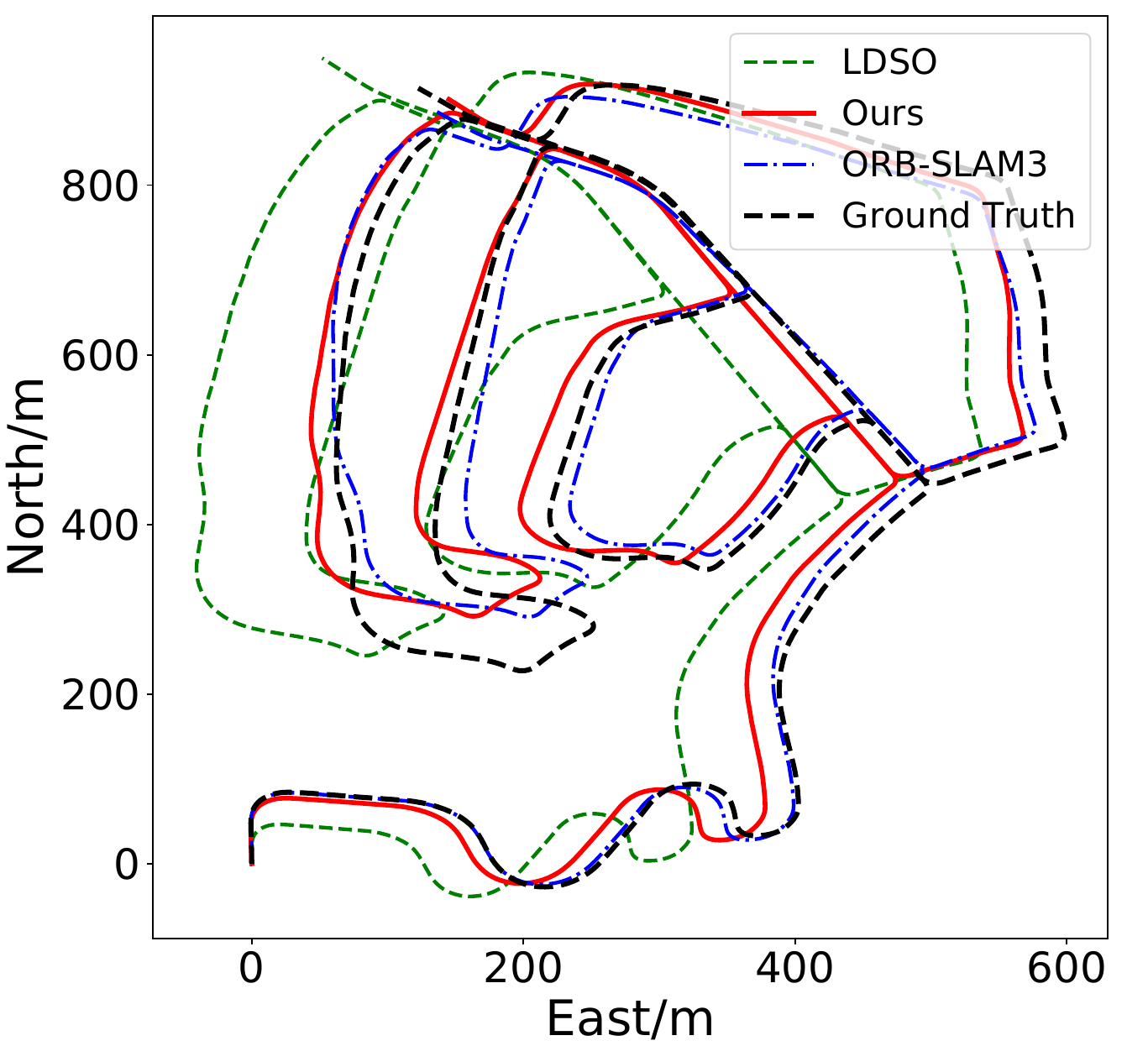}
 }
  \subfigure[Sequence 05]{ 
    \centering
    \includegraphics[width=0.315\linewidth]{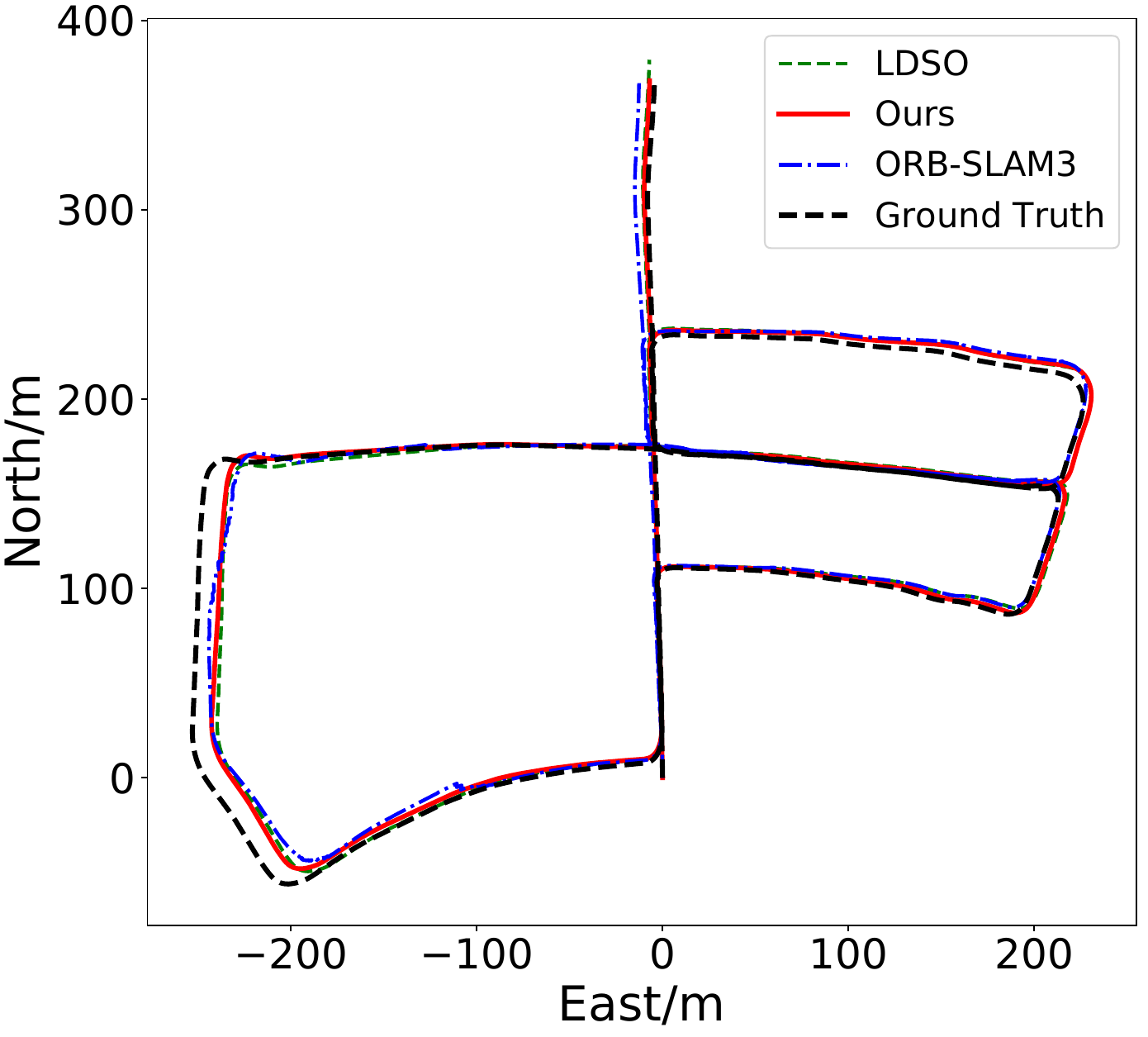}
 }

 \subfigure[Sequence 07]{ 
	\centering
	\includegraphics[width=0.315\linewidth]{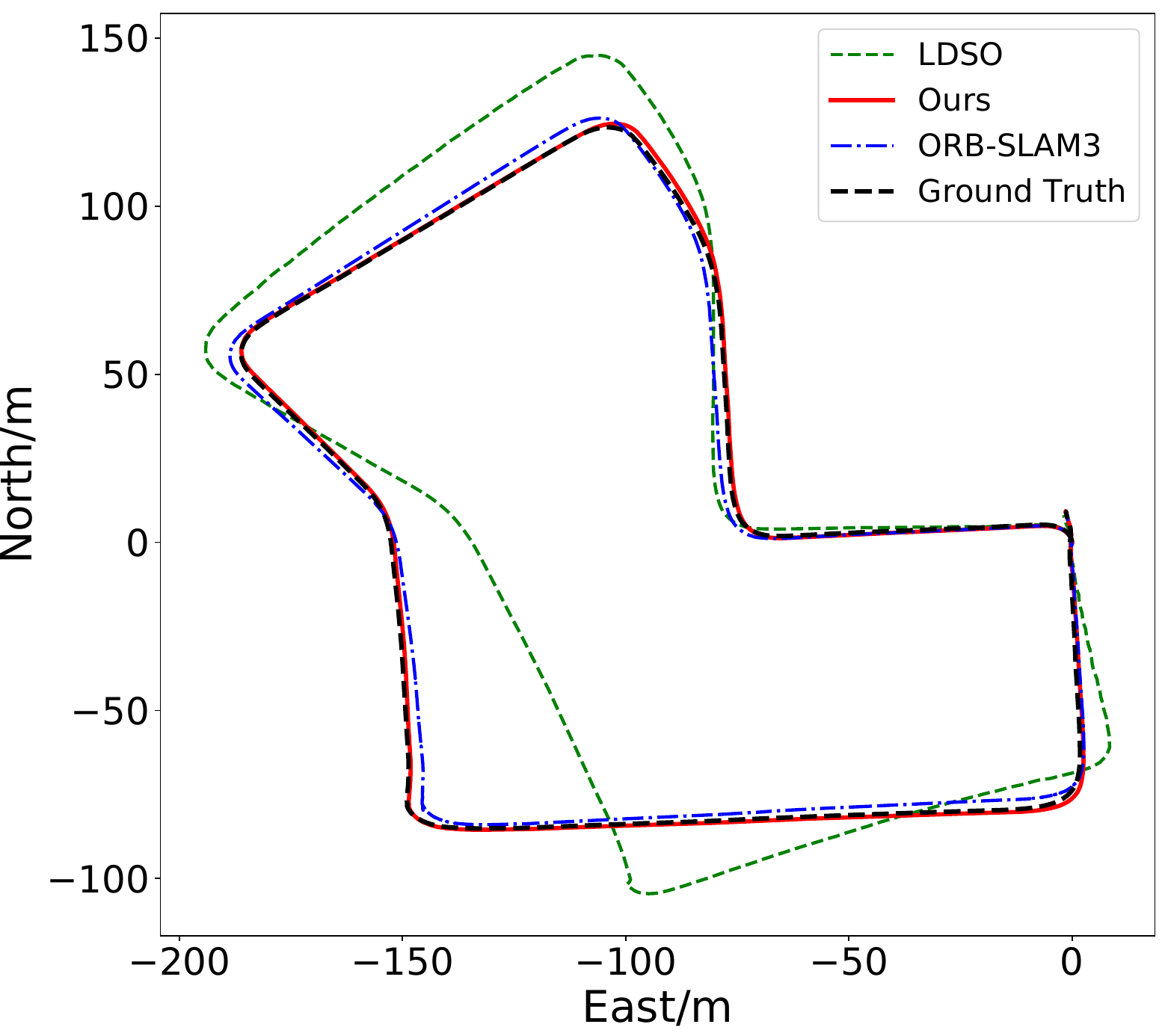}
 }
  \subfigure[Sequence 09]{ 
	\centering
	\includegraphics[width=0.315\linewidth]{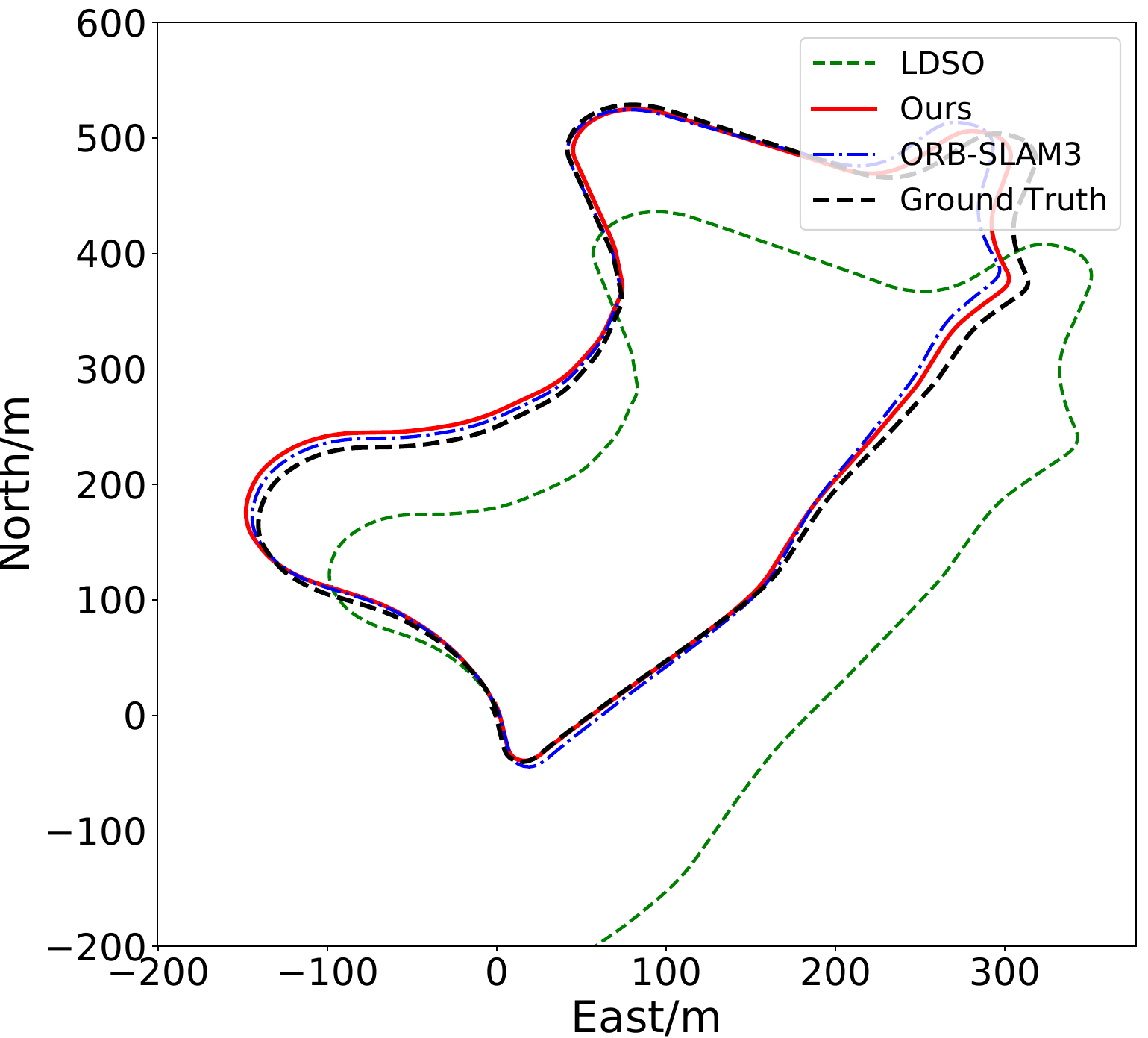}
 }
   \subfigure[Sequence 10]{ 
	\centering
	\includegraphics[width=0.315\linewidth]{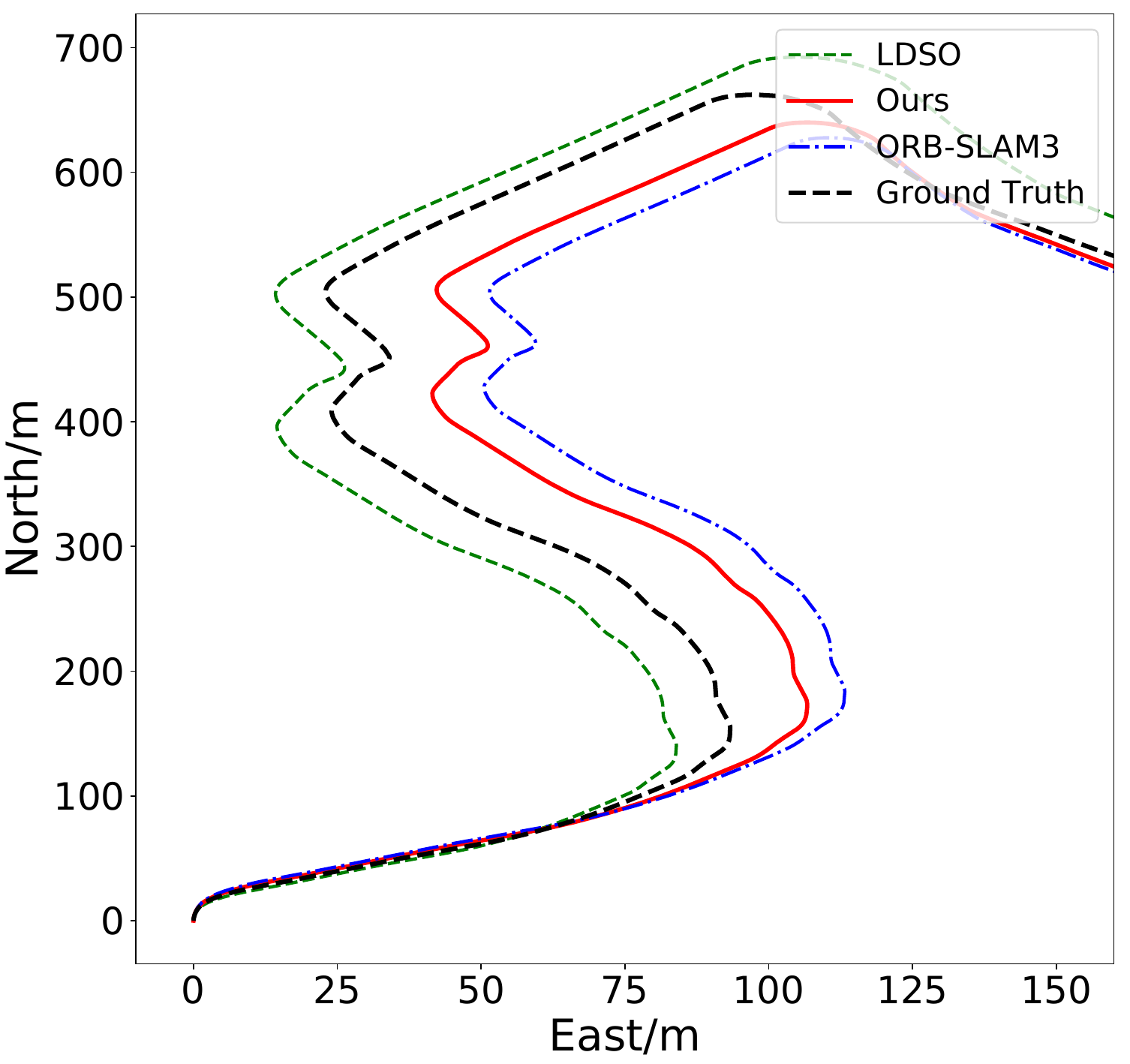}
 }
 \caption{The generated trajectories of our proposed DK-SLAM on the Sequence 00, 02, 05, 07, 09 and 10 of the KITTI dataset, comparing with LDSO and ORB-SLAM3.}  
\label{fig: depth kitti images}
\end{figure*}

\begin{figure*}[h]
    \centering
    \includegraphics[width=17cm]{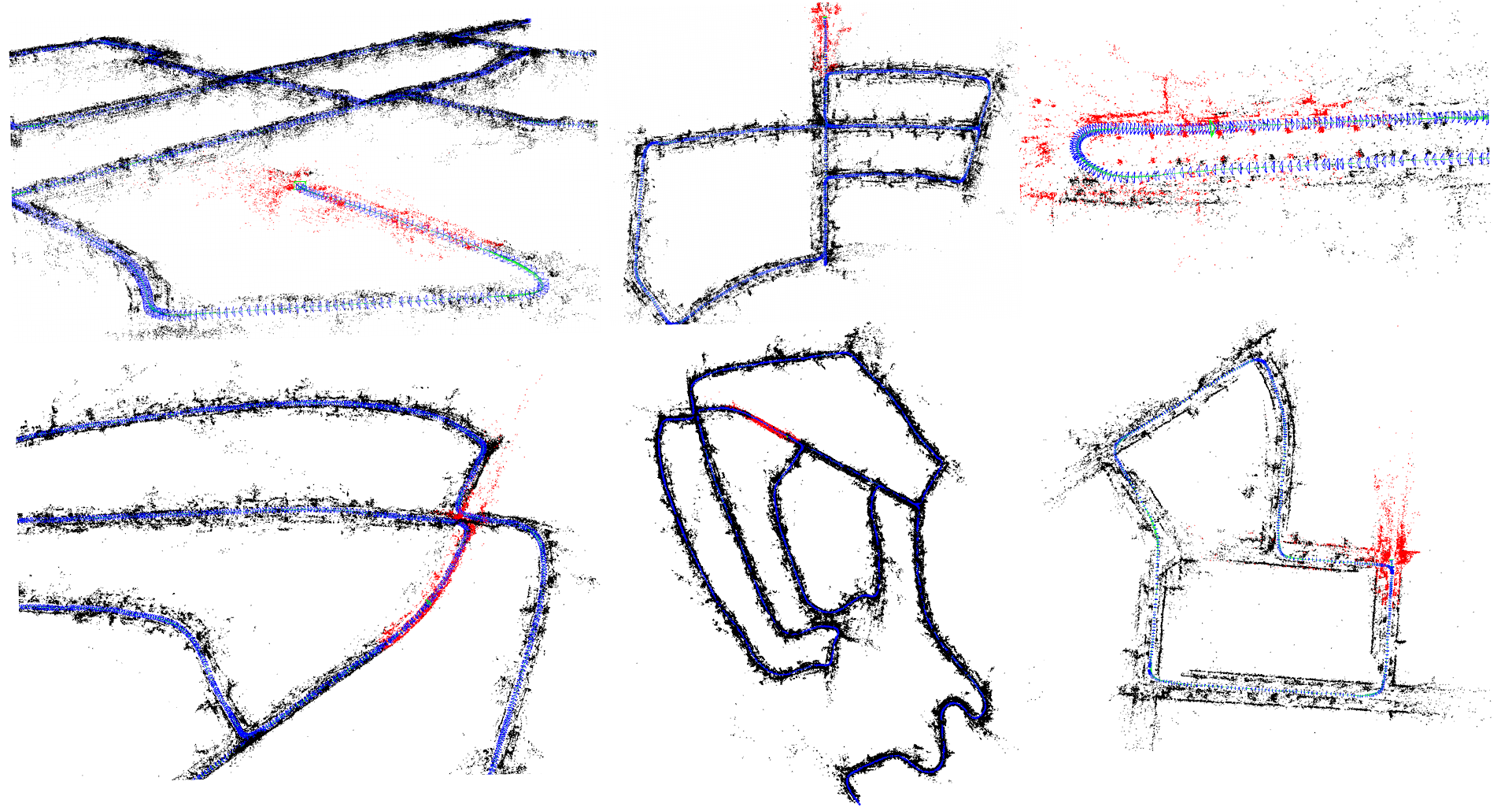}
    \caption{Mapping results generated by our proposed DK-SLAM system.}
    \label{mapping_overview}
\end{figure*}

\begin{equation}
\label{eq: Online Deep Bow}
\mathbf{\hat{d}_i} =\begin{cases}
1
  & \text{ if } \mathbf{d}_i\ge0  \\
0  & \text{ if } \mathbf{d}_i< 0
\end{cases} 
\end{equation}

\begin{table*}[h]
\centering
\caption{The pose evaluation on the KITTI 
dataset: our DK-SLAM system outperforms both representative traditional and learning-based monocular SLAM methods. Sequence 01 is excluded from the evaluation due to its inherent challenges for all monocular SLAM baselines, which prevent the generation of reasonable comparative results.}
\renewcommand\arraystretch{1.5}
\begin{tabular}{ccccccccccccccc}
\hline
Method              & Sensor                & Metric & 00 & 02 & 03& 04& 05 & 06&  07& 08& 09 & 10  & Avg &\\ \hline
\multirow{2}{*}{VISO-M} & \multirow{2}{*}{Mono} & $t_{rel}$ & 36.95 &  21.98&  16.14&  2.61& 17.20 & 7.91 & 20.00 & 39.78& 29.01 & 28.52 & 22.01 &\\ \cline{3-15} 
                    &                       & $r_{rel}$ & 2.42 &  1.22 & 2.67&  1.53& 3.52 &  1.83 & 5.30  &  1.99& 1.32 &  3.23 & 2.50 &\\ \hline
\multirow{2}{*}{LDSO} & \multirow{2}{*}{Mono} & $t_{rel}$ & 2.84 &   4.91 &2.75 & 1.58& 2.07 & \textbf{6.49} & 12.76 & 30.17 & 23.14 & 9.20 & 9.59 &\\ \cline{3-15} 
                    &                       & $r_{rel}$ & 0.36 &   0.74 & \textbf{0.16} & \textbf{0.19} & \textbf{0.24}& \textbf{0.16} &  5.13& 0.24 &  \textbf{0.21} &  \textbf{0.21} & 0.76 &\\ \hline
\multirow{2}{*}{ORB-SLAM3} & \multirow{2}{*}{Mono} & $t_{rel}$ & 2.80 &  6.04&  2.71&  2.22&  3.41&  7.07& 2.14&  10.35& 3.13 & 10.78 & 5.07 &\\ \cline{3-15} 
                    &                       & $r_{rel}$ & \textbf{0.3} &  0.38&  0.17&  0.24&   0.39 &  0.18& 0.42  & 0.31& 0.54 & 0.36 & 0.33 &\\ \hline
\multirow{2}{*}{LIFT-SLAM} & \multirow{2}{*}{Mono} & $t_{rel}$ & 3.18 &  8.73  &  \textbf{1.46}&  2.22& 6.09 & 12.24& 2.42 & 47.10& 19.91 & 9.72 & 11.31 &\\ \cline{3-15} 
                    &                       & $r_{rel}$ & 2.99 &  2.49  &  0.34&  0.48&  3.11 &  2.91& 4.02 & 2.02& 2.14 &  2.24 & 2.27 &\\ \hline
\multirow{2}{*}{DK-SLAM (Ours)} & \multirow{2}{*}{Mono} & $t_{rel}$ & \textbf{2.57} &  4.38  &  4.70&  \textbf{1.12}& \textbf{1.98} & 8.20 & \textbf{1.11}  & \textbf{7.61}& \textbf{2.97} & \textbf{7.03} & \textbf{4.17} &\\ \cline{3-15} 
                    &                       & $r_{rel}$ & 0.31 &  \textbf{0.27} & 0.17&  0.24& 0.26  & 0.17& 0.28 &  0.28&  0.29 &  0.23 & \textbf{0.25} &\\ \hline
\end{tabular}
\label{tb: Pose kitti}
\end{table*}

\subsubsection{Loop Node Detection}
As shown in the Figure \ref{online_tree}, the loop closure thread receives keyframes filtered from the local mapping thread, storing them in the inverted database to construct a Bag-of-Words (BoW) model. 
The BoW model utilizes the feature descriptors of the current frame $K_{c}$ to match descriptors stored in the database. Each keyframe is converted into a unique BoW vector for representation. We compute the similarity score between the current keyframe $K_{c}$ and the BoW vectors of inverted keyframes to identify the most similar candidate keyframe $K_{h}$. However, the candidate keyframe $K_{h}$ only considers differences in the numerical values of descriptors and neglects differences in descriptor positions, which may lead to mismatches.

Thus, we first employ the Grid-Based Motion Statistics (GMS) \cite{GMS} based brute force matching method to match deep keypoints between candidate closed-loop nodes. If the number of matches is below a threshold (i.e. 20 in our system), the candidate loop keyframe is classified as a mismatch. GMS operates on the principle that the number of keypoints near correctly matched feature points should surpass the number near incorrectly matched keypoints. Following a procedure similar to ORB-SLAM2\cite{ORB-SLAM2}, we compute the similarity transformation matrix $\mathbf{T}_\text{sim}$ between candidate closed-loop nodes through matched keypoints. Subsequently, $\mathbf{T}_\text{sim}$ is applied to match map points between the current keyframe $K_{c}$ and the candidate keyframe $K_{h}$. Further optimization of the similarity transformation matrix $\mathbf{T}_{sim}$ is conducted using the matched map points. If the number of inliers in the optimization function exceeds the threshold, the candidate keyframe $K_{h}$ is deemed a matched keyframe $K_{m}$.

\subsubsection{Global Map Correction via Loop-Closing}
Optimizing the similarity transformation matrix $\mathbf{T}_\text{sim}$ refines the pose of the current keyframe and its co-view related keyframes. Concurrently, we update the positions of co-observed map points across these keyframes. Next, we align the map points of the matched keyframe $K_{m}$ and its connected keyframes with those of the current keyframe $K_{c}$ and its connected keyframes. Updating the visibility graph between keyframes based on matched map points follows. Utilizing $K_{c}$, $K_{m}$, global map points, and the visibility graph, we construct and optimize the essential graph. Finally, a global bundle adjustment refines all keyframes and map point positions within the global map.

\section{Experiments}
\label{sec: experiments}
This section discusses the implementation details and evaluates the pose estimation performance of our proposed DK-SLAM on two widely adopted public datasets: the KITTI dataset, representing the car-driving scenario, and the EuRoC dataset, representing drone navigation. Moreover, we conducted an extensive ablation study to validate the effectiveness of our proposed key modules, including learned feature extraction, coarse-to-fine matching strategy, and online learning-based Bag-of-Words (BoW).

\subsection{Training Details and Datasets}
\subsubsection{Training Details} In MAML-based training, we set the batch size to 8, with 4 batches dedicated to support sets and 4 batches to query sets. Each minibatch is treated as a task, and 4 tasks are trained in the inner loop. The trained model calculates gradients on the query set, updating raw model parameters. The training spans 200K iterations, using the MS-COCO dataset \cite{COCO}.

\subsubsection{KITTI Odometry Dataset}
The KITTI odometry dataset \cite{KITTI} serves as a benchmark for self-driving scenarios, offering stereo RGB and grayscale images, along with Lidar and pose ground truth data. Grayscale images from sequences 00, 02, 03, 04, 05, 06, 07, 08, 09, and 10 are utilized for pose evaluation. Sequence 01 is excluded from the evaluation due to its inherent challenges for monocular SLAM systems.

\subsubsection{EuRoC MAV Dataset}
The EuRoC dataset \cite{EuROC} represents an indoor MAV flight dataset, encompassing stereo grayscale images, IMU data, and Leica Vicon's pose ground truth. Grayscale images are collected at a frequency of 20Hz. Due to space limitations, we focus on sequences MH01-05 for SLAM performance evaluation.

\subsection{Pose Evaluation in the Car-Driving Scenario} 
To evaluate the effectiveness of our SLAM system in the car-driving scenario, we conducted experiments using the KITTI dataset and employed the official evaluation metrics, computing root-mean-square error (RMSE) for both translation and rotation vectors. This evaluation spanned sequences ranging from 100m to 800m, providing an overall metric for pose accuracy. We conducted a comprehensive assessment of our SLAM system's tracking and loop closure capabilities in complex environments, focusing on Sequences 00, 02, 03, 04, 05, 06, 07, 08, 09, and 10 from the KITTI dataset. We excluded Sequence 01 from our evaluation due to its unique challenges for monocular SLAM systems, as a highway scene with few distinctive feature points, especially near the horizon, it led to significant scale degradation in monocular SLAM performance.

We benchmarked our DK-SLAM against several leading SLAM systems: LDSO \cite{LDSO}, ORB-SLAM3 \cite{ORB-SLAM3}, VISO-M \cite{VISO}, and LIFT-SLAM \cite{LIFT-SLAM}. Among these, VISO-M exhibited the least accuracy, particularly struggling in low-light conditions where its corner detection algorithms were unstable. LDSO, which includes a loop closure module, managed to reduce accumulated errors effectively in challenging scenarios. ORB-SLAM3, a sophisticated keypoint-based SLAM system that uses FAST corner detection and BRIEF descriptors, performed exceptionally well across most sequences as shown in Table \ref{tb: Pose kitti}. However, it showed a noticeable drop in performance on Sequence 10, primarily due to a scarcity of reliable corner points, which resulted in unstable tracking and adversely affected overall pose estimation accuracy. The reliance on BRIEF, which describes local photometric information around corners, made ORB-SLAM3 particularly sensitive to lighting changes, thereby impacting feature tracking performance. In environments with significant lighting variations, the inability to track reliable feature points at the front-end increased positioning errors. Although the back-end loop closure could correct some accumulated errors, it struggled to improve overall accuracy. In contrast, LIFT-SLAM, which depends on learning-based local features, generally underperformed across most sequences. This outcome suggests that while learning-based features have potential, they do not consistently outperform traditional feature-based methods. ORB-SLAM3’s attention to scale, direction, and a traditional SLAM system for achieving even keypoint distribution gave it a competitive advantage, helping to filter out mismatched features at the front-end.


Our DK-SLAM system demonstrated exceptional performance across most KITTI sequences. Specifically, DK-SLAM shows a substantial improvement over traditional monocular ORB-SLAM3, achieving approximately 17.7\% better translation accuracy and 24.2\% higher rotation accuracy on the KITTI dataset. When compared to LIFT-SLAM, a leading SLAM system based on learned features, DK-SLAM surpasses it by nearly 2.7 times in translation accuracy and 9 times in rotation accuracy. This significant enhancement in performance is attributed to DK-SLAM's use of meta learning based feature extraction and a coarse-to-fine matching strategy, which focuses on keypoint-surrounding patches and optimizes pose estimation through precise 3D-2D matching relationships. Additionally, our system's online training capability builds a learning-based Bag-of-Words (BoW) model using previously acquired data, efficiently constraining the BoW’s feature description space and improving loop scene detection. This approach enables the loop closure module to accurately correct cumulative errors. Moreover, as illustrated in Figure \ref{mapping_overview}, DK-SLAM excels in mapping performance, providing detailed geometric insights. The incorporation of MAML-based feature training enhances the generalization capability of the feature extractor, allowing it to capture more comprehensive scene details. This combination of strategies ensures that DK-SLAM offers robust and accurate performance in varied and complex environments.

\begin{figure*}[h]
	\centering
 \subfigure[Sequence MH01]{ 
	\centering
	\includegraphics[width=0.315\linewidth]{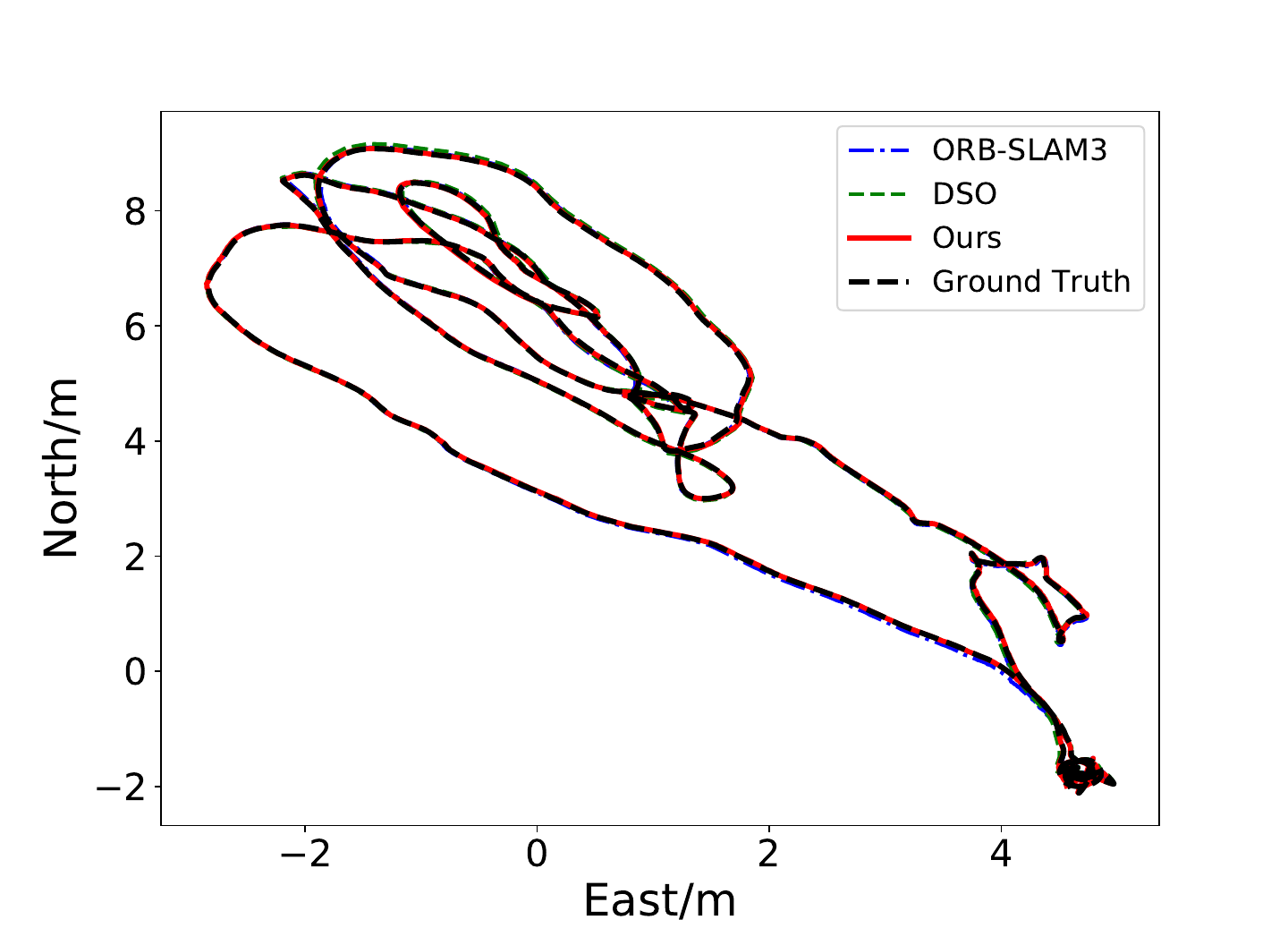}
 }
  \subfigure[Sequence MH02]{ 
    \centering
    \includegraphics[width=0.315\linewidth]{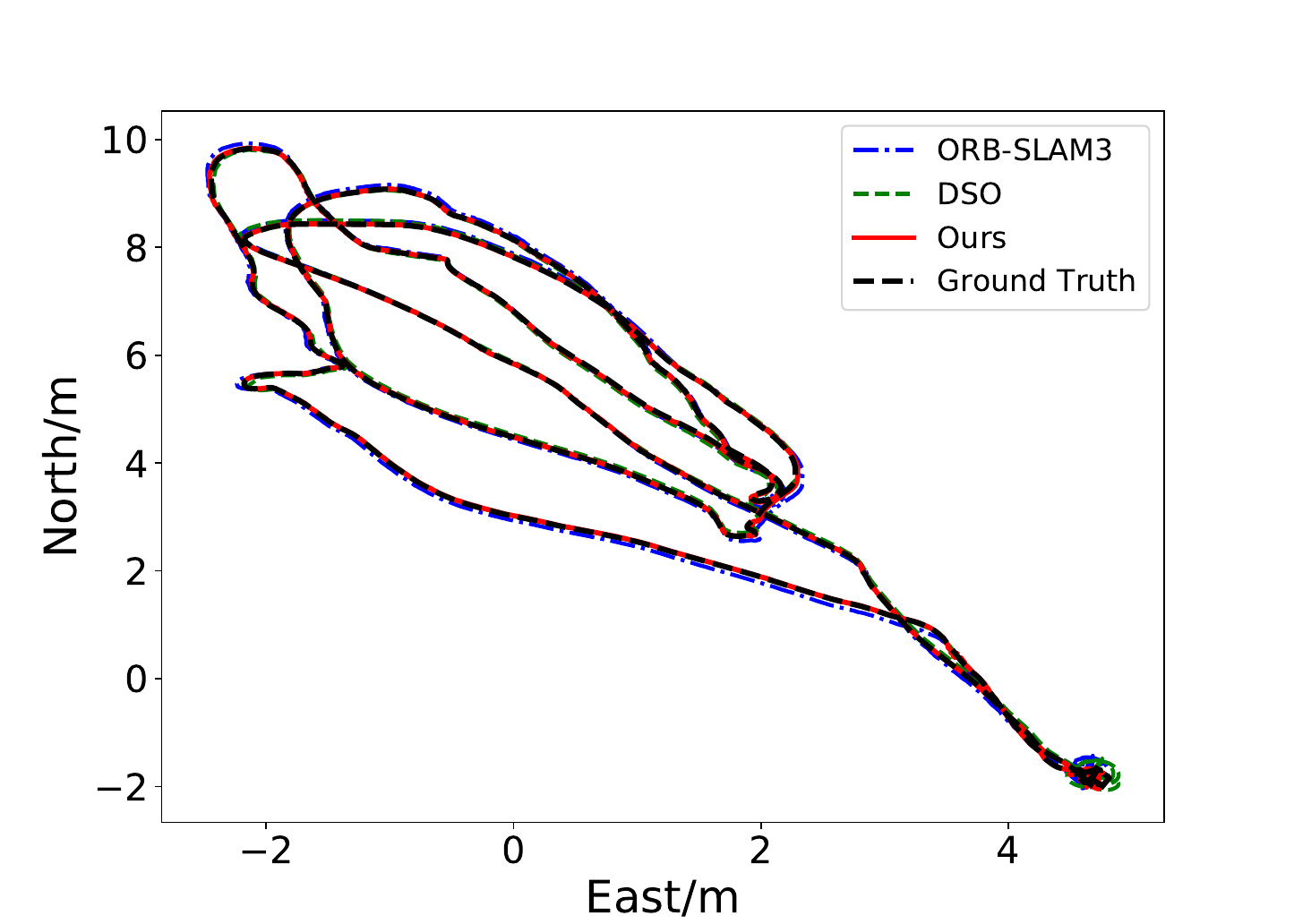}
 }
  \subfigure[Sequence MH03]{ 
    \centering
    \includegraphics[width=0.315\linewidth]{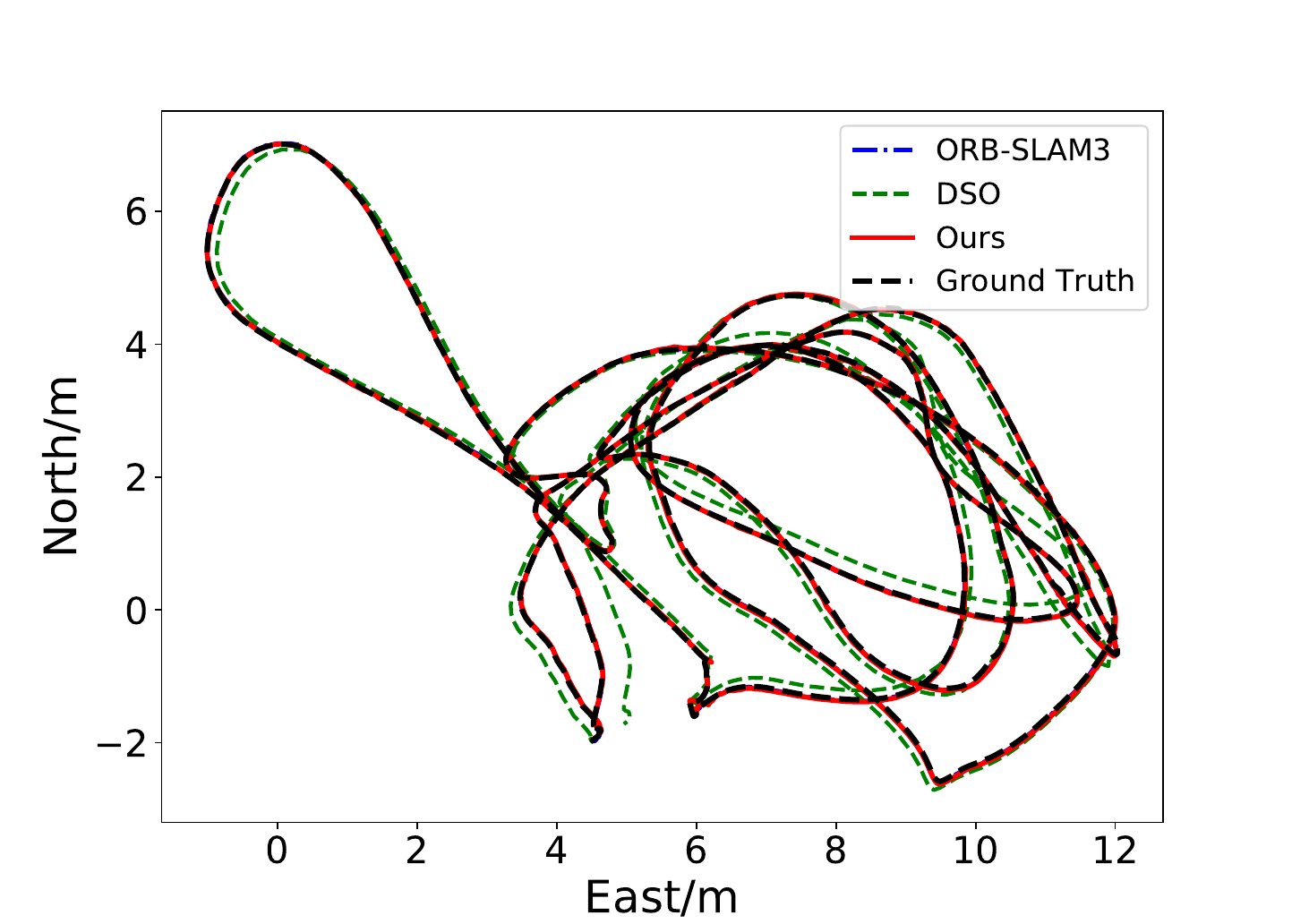}
 }

 \subfigure[Sequence MH04]{ 
	\centering
	\includegraphics[width=0.315\linewidth]{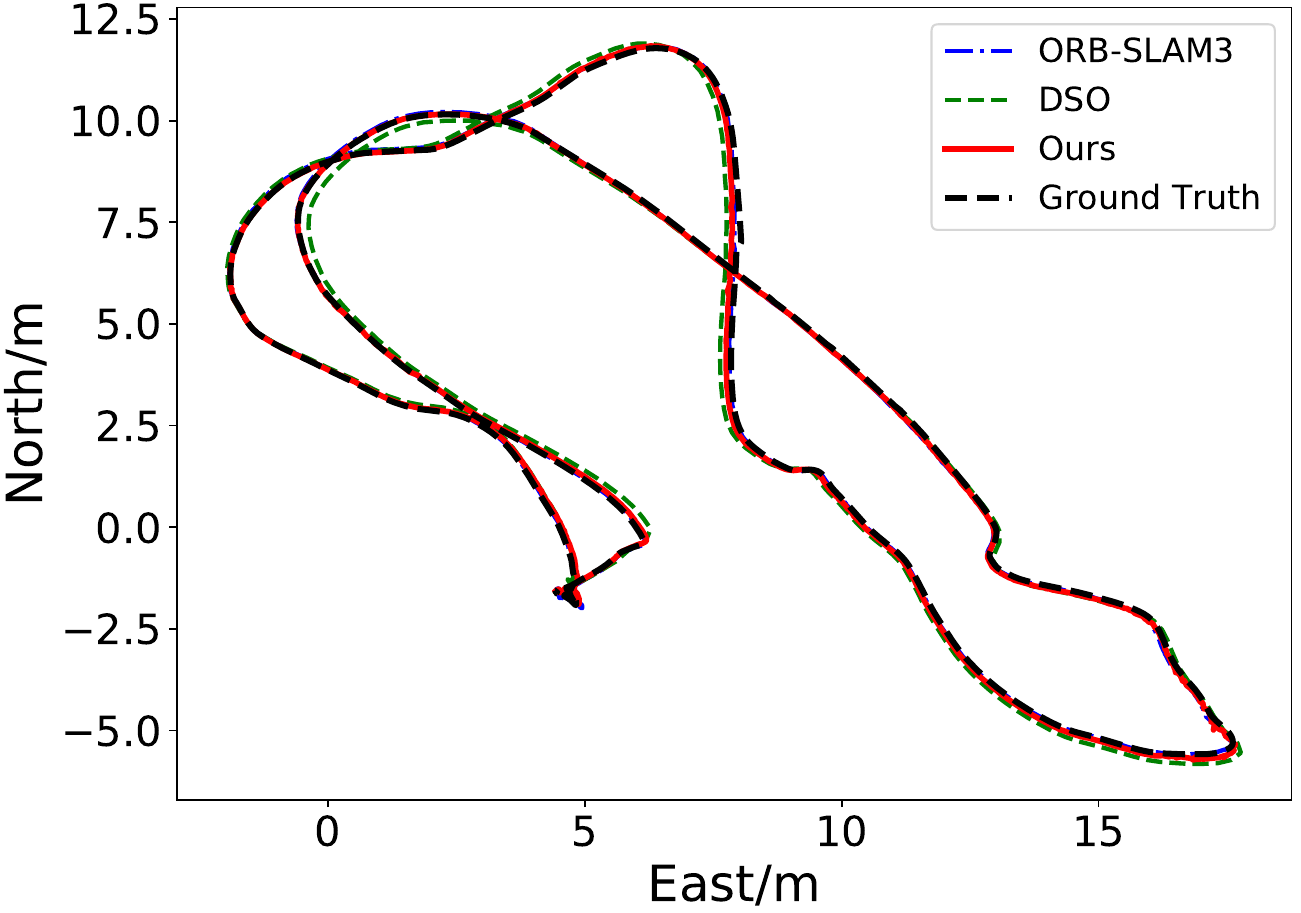}
 }
  \subfigure[Sequence MH05]{ 
	\centering
	\includegraphics[width=0.315\linewidth]{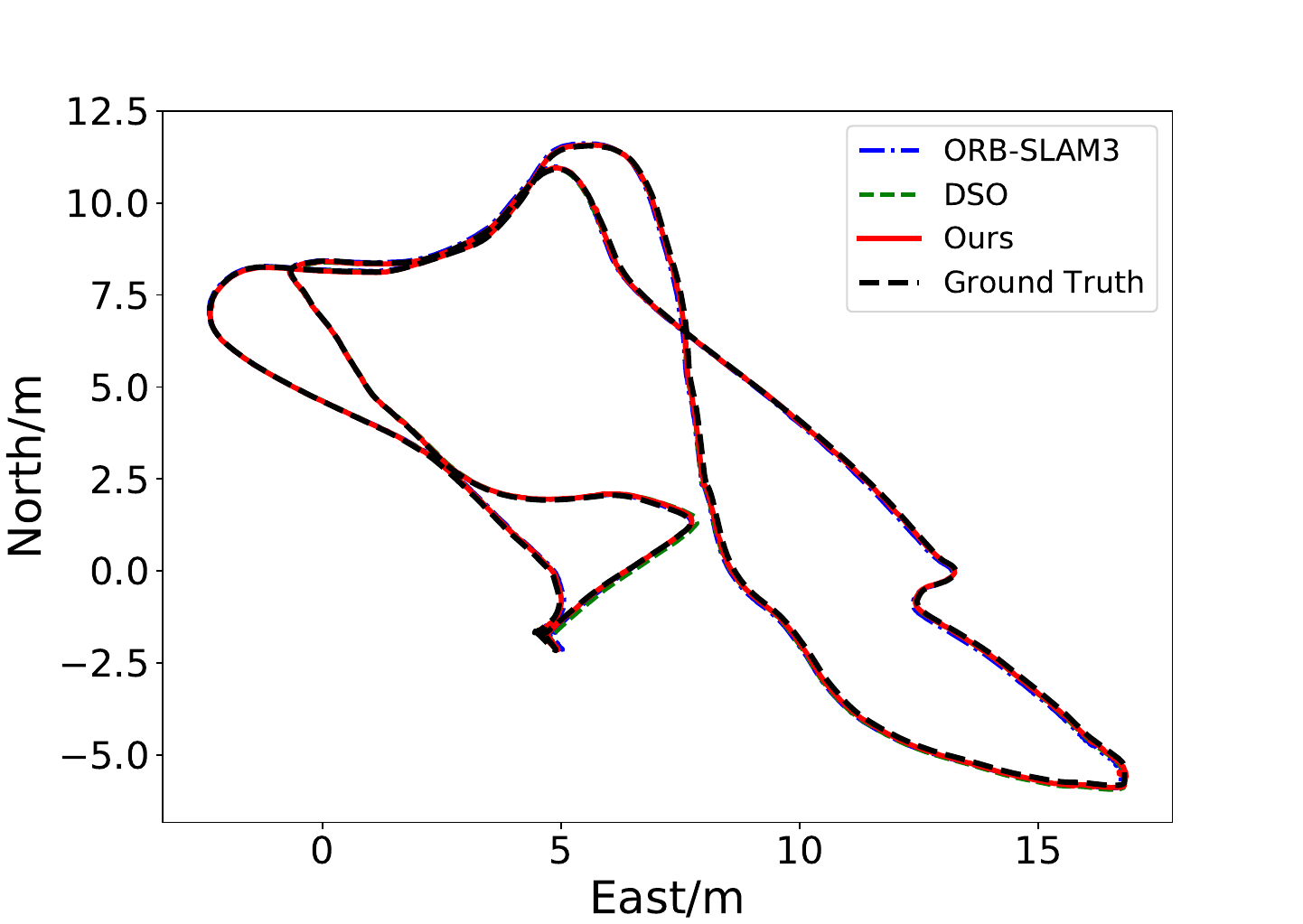}
 }
 \caption{The generated trajectories of our proposed DK-SLAM on the Sequence MH01, MH02, MH03, MH04 and MH05 of the EuRoC dataset, comparing with DSO and ORB-SLAM3.}  
\label{fig: euroc images}
\end{figure*}

\begin{table}[h!]
\small
\centering
\caption{Comparison of absolute translation errors (in meters) between the proposed DK-SLAM and other baselines on the EuRoC dataset. Scaling with the ground truth is necessary for evaluation due to the absence of absolute scale in monocular visual SLAM methods, as noted in the table. "-" indicates performance failure.}
\setlength{\tabcolsep}{1.2mm}{
\renewcommand{\arraystretch}{1.5}
\begin{tabular}{ccccccc}
\hline
\multicolumn{1}{c}{Method}  &\multicolumn{1}{c}{MH01}&\multicolumn{1}{c}{MH02} & \multicolumn{1}{c}{MH03} & \multicolumn{1}{c}{MH04} &\multicolumn{1}{c}{MH05}   & \multicolumn{1}{c}{Avg}\\ \hline
DSO     & 0.046 & 0.046 & 0.172         & 3.810   &  0.110    & 0.8368   \\
SVO       & 0.100 & 0.120        & 0.410      & 0.430    & 0.300           &  0.2720\\
DSM       & 0.039 & 0.036        & 0.055      & \textbf{0.057}    & 0.067        & 0.0508  \\
ORB-SLAM3  & 0.016 & 0.027    & 0.028     & 0.138   & 0.072       &  0.0562  \\ 
LIFT-SLAM   & 0.044 & 0.053   & 0.049   & -  & -    & - \\
\hline 
DK-SLAM       & \textbf{0.013} &  \textbf{0.013}   & \textbf{0.027}  & 0.077  & \textbf{0.055}   & \textbf{0.0370}\\ 
\hline
\end{tabular}}
\label{tb: pose Euroc}
\end{table}

\begin{table*}[h]
\centering
\caption{The ablation study into the key modules in our DK-SLAM system.}
\renewcommand\arraystretch{1.5}
\begin{tabular}{cccccccccccccccc}
\hline
Method              & Meta Learning & Two-Stage Tracking             & Metric & 00 & 02 & 03& 04& 05 & 06& 07& 08& 09 & 10  & Avg &\\ \hline
\multirow{2}{*}{Ours1} & \multirow{2}{*}{{$\checkmark$}}  &\multirow{2}{*}{} & $t_{rel}$ & 2.88  & -  & \textbf{4.29} &1.24 & 2.80 &\textbf{7.64} & 1.76 & -& \textbf{2.70} & - & - &\\ \cline{5-16} 
                       &     &              & $r_{rel}$ & 0.41  & -  &  \textbf{0.16}& 0.31 & 0.35 & 0.18 & 0.64 & -& 0.24 &  - & - &\\ \hline
\multirow{2}{*}{Ours2} &\multirow{2}{*}{}&\multirow{2}{*}{{$\checkmark$}}& $t_{rel}$ & 2.92  & 4.42  & 4.54& 2.01& 2.48& 7.97 & 2.11 & 8.85 & 4.03 & 8.48 & 4.78 &\\ \cline{5-16} 
                        &     &              & $r_{rel}$ & 0.38  & 0.34 &  0.21& \textbf{0.16} & 0.34 & 0.18 & 0.46& \textbf{0.26} & \textbf{0.20} &  0.23 & 0.28 &\\ \hline
\multirow{2}{*}{Ours3} &\multirow{2}{*}{$\checkmark$}&\multirow{2}{*}{$\checkmark$}& $t_{rel}$ & \textbf{2.57}  & \textbf{4.38} & 4.70& \textbf{1.12}& \textbf{1.98} &  8.20& \textbf{1.11} & \textbf{7.61}& 2.97 & \textbf{7.03} & \textbf{4.17} &\\ \cline{5-16} 
                    &             &         & $r_{rel}$ & \textbf{0.31}  & \textbf{0.27} & 0.17& 0.24& \textbf{0.26}  &\textbf{ 0.17}& \textbf{0.28} & 0.28& 0.29 &  \textbf{0.23} & \textbf{0.25} &\\ \hline
\end{tabular}
\label{tb: ablation}
\end{table*}

\subsection{Pose Evaluation in the UAV Scenario}
As shown in Table \ref{tb: pose Euroc} and Figure \ref{fig: euroc images}, we come to validate our proposed DK-SLAM on the EuRoC dataset. Following the previous research, we address the scale ambiguity with umeyama alignment \cite{Umeyama} for monocular vision methods. Comparisons include ORB-SLAM3 \cite{ORB-SLAM3}, DSO \cite{DSO}, DSM \cite{DSM}, SVO \cite{SVO}, and LIFE-SLAM \cite{LIFT-SLAM}. Results, derived from \cite{DROID} for ORB-SLAM3, DSO, DSM, and SVO, and \cite{LIFT-SLAM} for LIFE-SLAM. We utilize the absolute translation error as the evaluation metric. 

The evaluation results reveal that the learned feature based SLAM, i.e. LIFE-SLAM, often fails in drone scenarios, as it relies on LIFT as a binary feature extractor and uses the Hamming distance between keypoints in consecutive frames for matching. In addition, LIFE-SLAM does not consider the structural relationships between keypoints, leading to a significant number of mismatches and subsequent feature tracking failures. ORB-SLAM3 \cite{ORB-SLAM3} employs an offline-trained Bag-of-Words (DBow3) for loop closure. Due to the sensitivity of its handcrafted descriptors to lighting variations, ORB-SLAM3 struggles with lower loop detection accuracy in complex lighting conditions, which diminishes the overall performance of the system.

In contrast, our DK-SLAM demonstrates outstanding localization performance. It uses a two-stage tracking strategy that precisely locates feature points and minimizes incorrect matches. The learned features in DK-SLAM are robust, performing well even in low-light conditions and supporting stable feature tracking. Moreover, the online learning-based deep Bag-of-Words model excels in loop detection. This is particularly evident in the MH02 sequence, where DK-SLAM surpasses ORB-SLAM3. This superior performance is due to the use of learned features for constructing the Bag-of-Words model, which provides detailed scene differentiation, and the online training capability, which allows for rapid adaptation to changing environments.

\begin{figure*}[h]
    \centering
    \includegraphics[width=12.0cm]{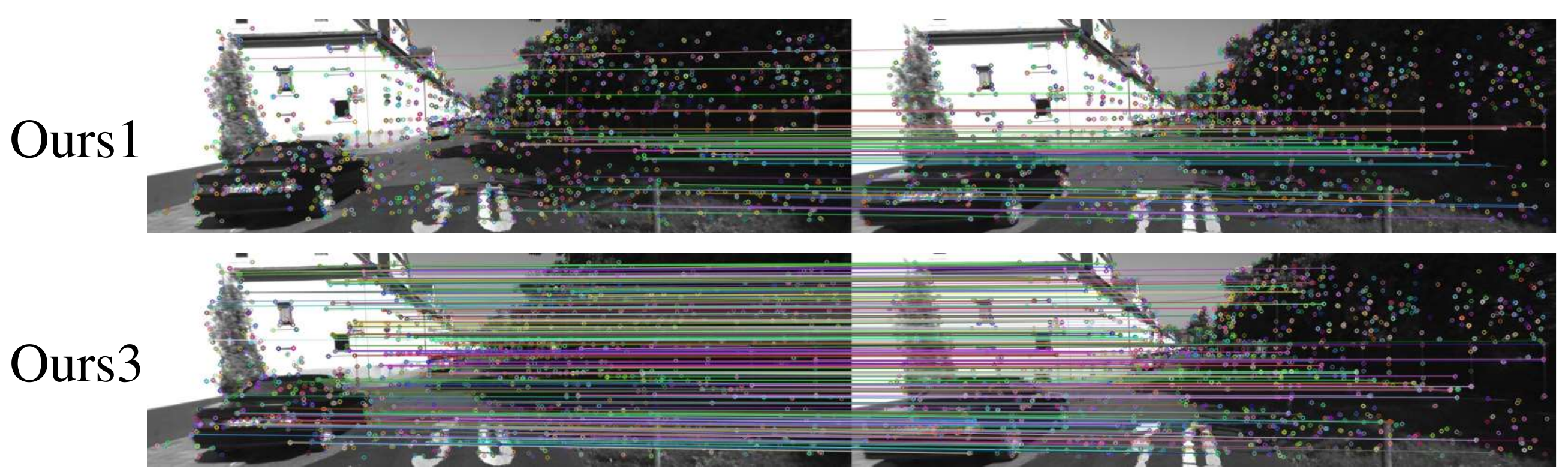}
    \caption{The samples of keypoint detection and matching. Up: matching without two-stage strategy (Ours1). Bottom: matching with two-stage strategy (Ours3).}
    \label{match_semi}
\end{figure*}

\begin{table}[h!]
\small
\centering
\caption{{The average number of matching points for ORB-SLAM3, and DK-SLAM without (Ours2) or with feature meta learning (Ours3).}}
\renewcommand{\arraystretch}{1.5}
\begin{tabular}{cccc}
\hline
\multicolumn{1}{c}{Seq}  &\multicolumn{1}{c}{ORB-SLAM3}&\multicolumn{1}{c}{Ours2} & \multicolumn{1}{c}{Ours3} \\ \hline
K00     & 228.3 & 352.6 & \textbf{376.7}          \\
K05       & 219.5 & 371.2        & \textbf{383.1}     \\
K09       & 150.4 & 258.0        & \textbf{265.3}      \\
K10       & 171.4 & 306.0        & \textbf{323.1}      \\
\hline
\end{tabular}
\label{mean_match_num}
\end{table}

\begin{figure*}[h]
    \centering
    \includegraphics[width=12.0cm]{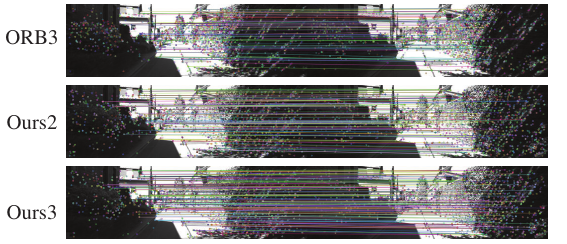}
    \caption{The samples of keypoints detection and matching. From top to bottom: ORB-SLAM3 matching (ORB3), DK-SLAM without (Ours2) or with keypoint meta learning (Ours3).}
    \label{match_three}
\end{figure*}

\subsection{Ablation Study}
Table \ref{tb: ablation} summarizes the results of the ablation study into the key modules within DK-SLAM system, assessing the impact of our MAML-based deep keypoint meta learning and coarse-to-fine keypoint tracking. For a fair comparison, "Ours1", "Ours2" and "Ours3" all use a keypoint search radius of 7.

\textbf{1) The analysis of keypoint meta learning module:} 
the module's effectiveness is evident when comparing "Ours2" and "Ours3." "Ours2," which employs the original SuperPoint strategy without our keypoint meta-learning approach, performs slightly worse than "Ours3," which integrates MAML-based keypoint meta-learning. In "Ours3," MAML is applied to train the local feature extractor by iteratively enhancing generalization on support and query sets, resulting in a robust and adaptable detector. The batch size of "Ours2" and "Ours3" is set as 8. Both models were trained 150K iterations. In "Ours3", we split every training batch to 4 mini-batch support set and 4 mini-batch query set. As mentioned in algorithm \ref{alg:Framwork}, the number of update iterations in inner loop is 4. 

Table \ref{mean_match_num} presents the average number of matching points for ORB-SLAM3, Ours2, and Ours3 on Sequence 00, 05, 09, and 10 of the KITTI dataset. Setting the total number of keypoints to 3000 and pyramid level to 3, MAML-based training significantly improves the feature detection performance of SuperPoints, allowing them to acquire more robust features and enhancing SLAM accuracy.

\textbf{2) The analysis of coarse-to-fine tracking module:}
in "Ours1," lacking a coarse-to-fine tracking strategy, reliance on a constant velocity motion model for keypoint matching leads to errors, resulting in inaccurate correspondence and tracking failure.  
Figure \ref{match_semi} shows a failure case for "Ours1" in accurate feature matching due to uniform motion model challenges. In contrast, "Ours3" uses a two-stage strategy for stable tracking, combining a semi-direct method for coarse pose estimation and refined feature matching. 

The enhanced SLAM performance attributed to proposed keypoint meta learning and two-stage matching strategy can also be observed in feature matching results, as depicted in Figure \ref{match_three}. Using SuperPoint for matching increases the number of matches, addressing ORB-SLAM3's limitations in texture-less areas. ORB-SLAM3's descriptors, describing lighting changes around corners, are susceptible to environmental variations. In contrast, the original learning-based local feature methods such as SuperPoint capture robust deep features unaffected by lighting changes, ensuring stable tracking across consecutive frames.

\begin{figure}[h]
    \centering
    \includegraphics[width=9cm]{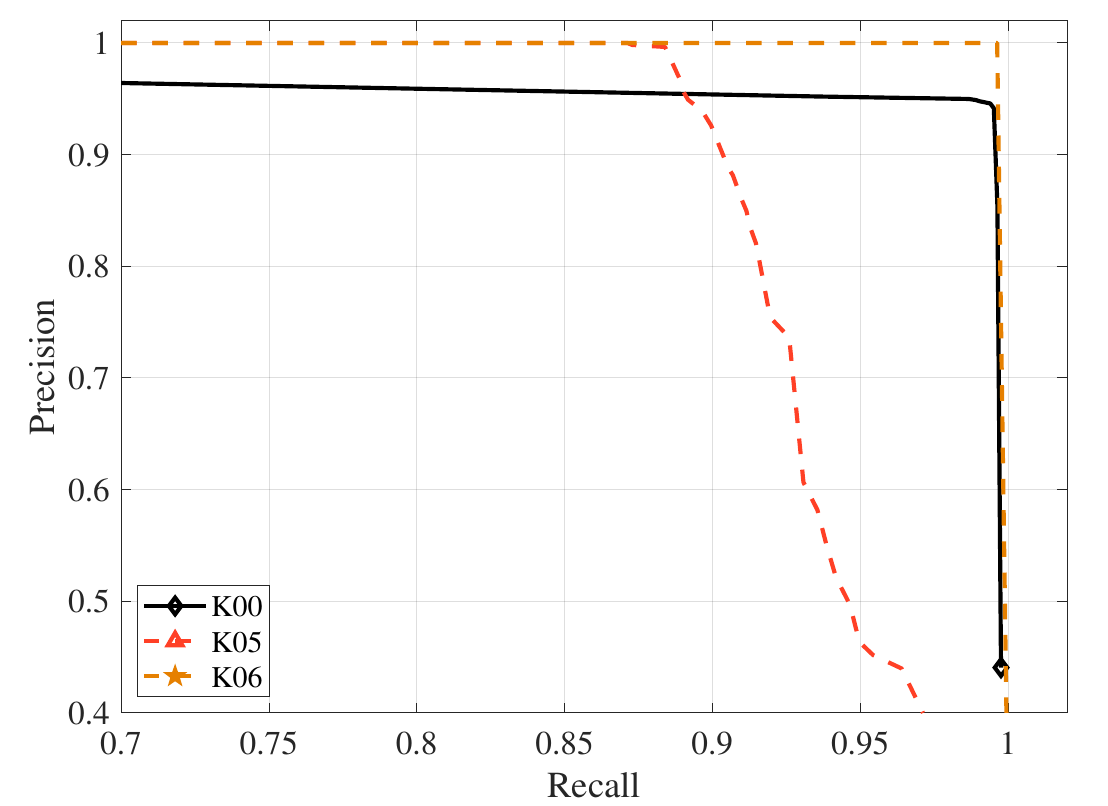}
    \caption{Precision-Recall curves depicting the performance of the Bag-of-Words (BoW) approach in the proposed DK-SLAM on Sequences 00, 05, and 06 of the KITTI dataset.}
    \label{PR_curves}
\end{figure}

\begin{table}[h!]
\small
\centering
\caption{
Performance comparison of loop-closing with the maximum recalls across various methods, all achieving 100\% precision. The numerical data presented in this table is sourced from \cite{iBoW,Robust_super_graph}, and \cite{ITSC}.}
\setlength{\tabcolsep}{1.2mm}{
\renewcommand{\arraystretch}{1.5}
\begin{tabular}{ccccccc}
\hline
\multicolumn{1}{c}{Seq}  &\multicolumn{1}{c}{FAB-MAP2}&\multicolumn{1}{c}{Emilio} & \multicolumn{1}{c}{Milford} & \multicolumn{1}{c}{seqSLAM} &\multicolumn{1}{c}{iBoW}   & \multicolumn{1}{c}{Ours}\\ \hline
K00     & 0.49 & 0.90 & 0.67         & 0.67   &  0.77    & \textbf{0.98}   \\
K05       & 0.32 & 0.76        & 0.41      & 0.36    & 0.26           &  \textbf{0.87}\\
K06       & 0.55 & 0.95        & 0.65      & 0.65    & 0.96        & \textbf{1}  \\
\hline
\end{tabular}}
\label{tb: Bow_maxRecall}
\end{table}

\textbf{3) The analysis of loop-closing module:} 
we further perform ablation studies on the proposed loop closure method to assess its performance. Precision-Recall metrics on KITTI 00, 05, and 06 sequences are illustrated in Figure \ref{PR_curves}, with additional evaluation at 100\% precision. The quantitative results, presented in Table \ref{tb: Bow_maxRecall}, reveal our deep BoW's superior recall rate compared to traditional BoWs. Notably, in the KITTI 06 sequence, our BoW achieves a remarkable 100\% recall rate. In contrast, the recall rate of the iBoW, belonging to the traditional BoW category, is significantly lower, suggesting that handcrafted descriptors might struggle to accurately identify loop nodes in a sequence. The learned local descriptor captures high-level information, enhancing robustness across diverse environments. This feature stability, unaffected by lighting changes, results in superior loop detection performance.

\section{Conclusion}
\label{sec: conclusion}

This work presents DK-SLAM, a monocular visual SLAM with deep keypoint meta learning, a coarse-to-fine matching strategy, and an online binary learned feature BoW. To enhance generalization, we employ MAML for local feature training, yielding a feature extractor with robust capabilities in unseen scenarios. Coarse relative poses between frames are estimated via a semi-direct method for accurate feature point matching. An online binary learned feature BoW corrects SLAM accumulation errors. Our proposed DK-SLAM outperforms representative baselines, e.g. ORB-SLAM3, on the KITTI and EuRoC datasets.  However, the GPU-based online front-end faces efficiency challenges during information transmission to the CPU-based back-end. Future work will explore knowledge distillation for neural network parameters compression and further SLAM framework modifications to enhance efficiency.

\bibliographystyle{IEEEtran}
\bibliography{ref}

\par\noindent 
\parbox[t]{\linewidth}{
\noindent\parpic{\includegraphics[height=1.5in,width=1in,clip,keepaspectratio]{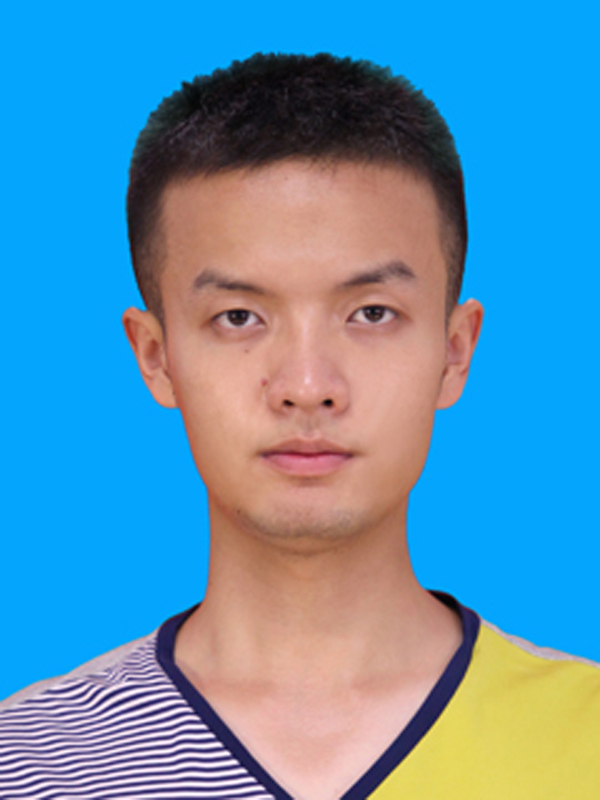}}
\noindent {\bf Hao Qu} is a Ph.D. student at College of Intelligence Science and Technology, National University of Defense Technology (China). Before that, he obtained his M.Eng. degree at National University of Defense Technology (China), and B.Eng.degree at National University of Defense Technology (China). His research interest lies in robotics, computer vision and navigation systems.
}
\vspace{2\baselineskip}

\par\noindent 
\parbox[t]{\linewidth}{
\noindent\parpic{\includegraphics[height=1.5in,width=1in,clip,keepaspectratio]{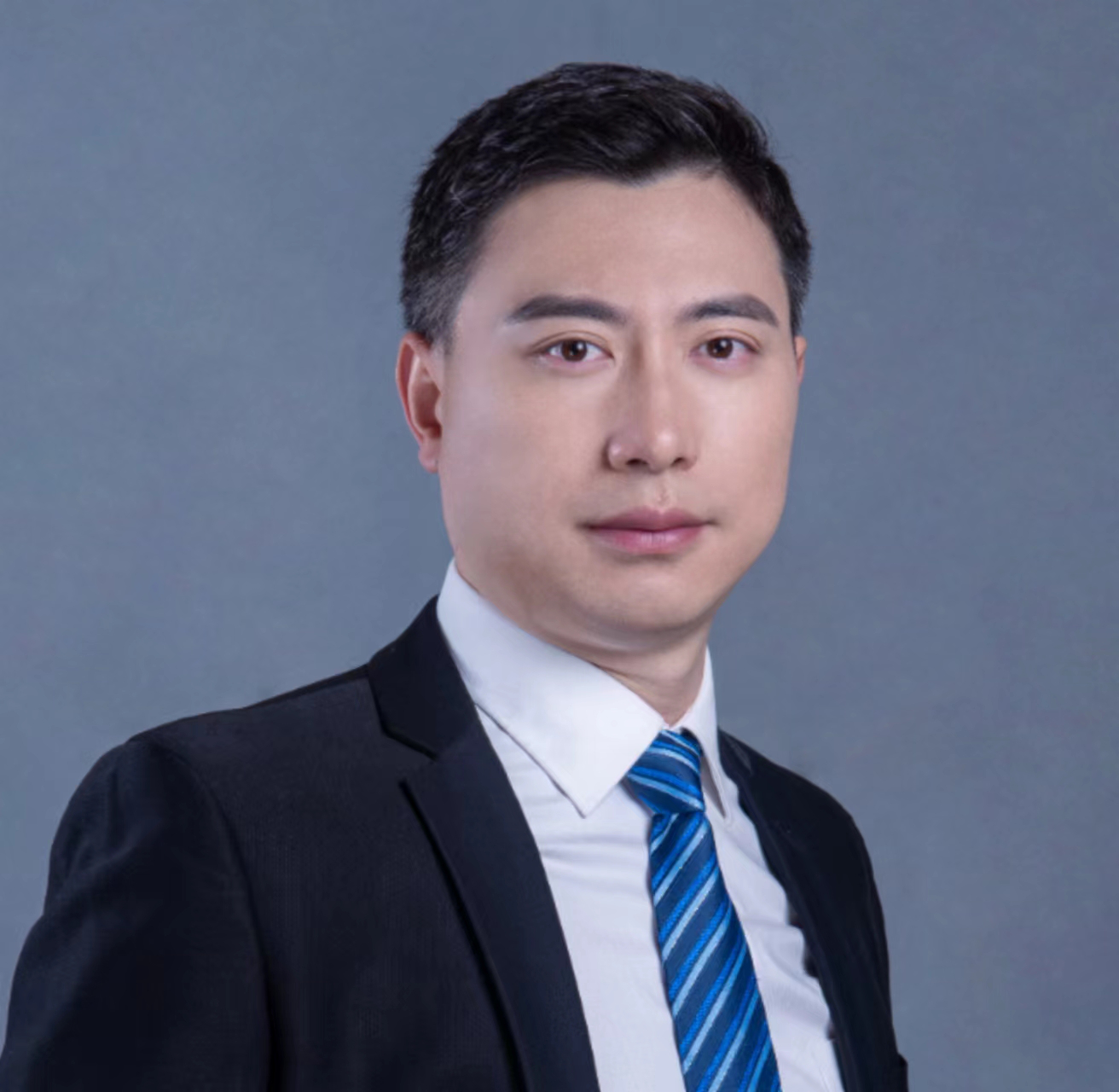}}
\noindent {\bf Lilian Zhang} received the B.S. degree from the School of Mechatronics and Automation,  National University of Defense Technology, China, in 2007, and the Ph.D. degree from  the Institute of Computer Science, University of Kiel, Germany, in 2013. He is currently an Associate Professor with the National University of Defense Technology.  His current  research interests include robotic vision and bionic navigation. 
}
\vspace{2\baselineskip}

\par\noindent 
\parbox[t]{\linewidth}{
\noindent\parpic{\includegraphics[height=1.5in,width=1in,clip,keepaspectratio]{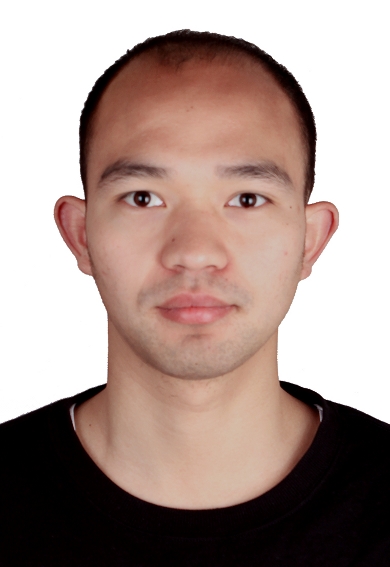}}
\noindent {\bf Mao Jun} received the B.S. and Ph.D degree from the School of Mechatronics and Automation,  National University of Defense Technology, China. He is currently an Associate Professor with the National University of Defense Technology.  His current  research interests include robotic vision and bionic navigation. 
}
\vspace{2\baselineskip}

\par\noindent 
\parbox[t]{\linewidth}{
\noindent\parpic{\includegraphics[height=1.5in,width=1in,clip,keepaspectratio]{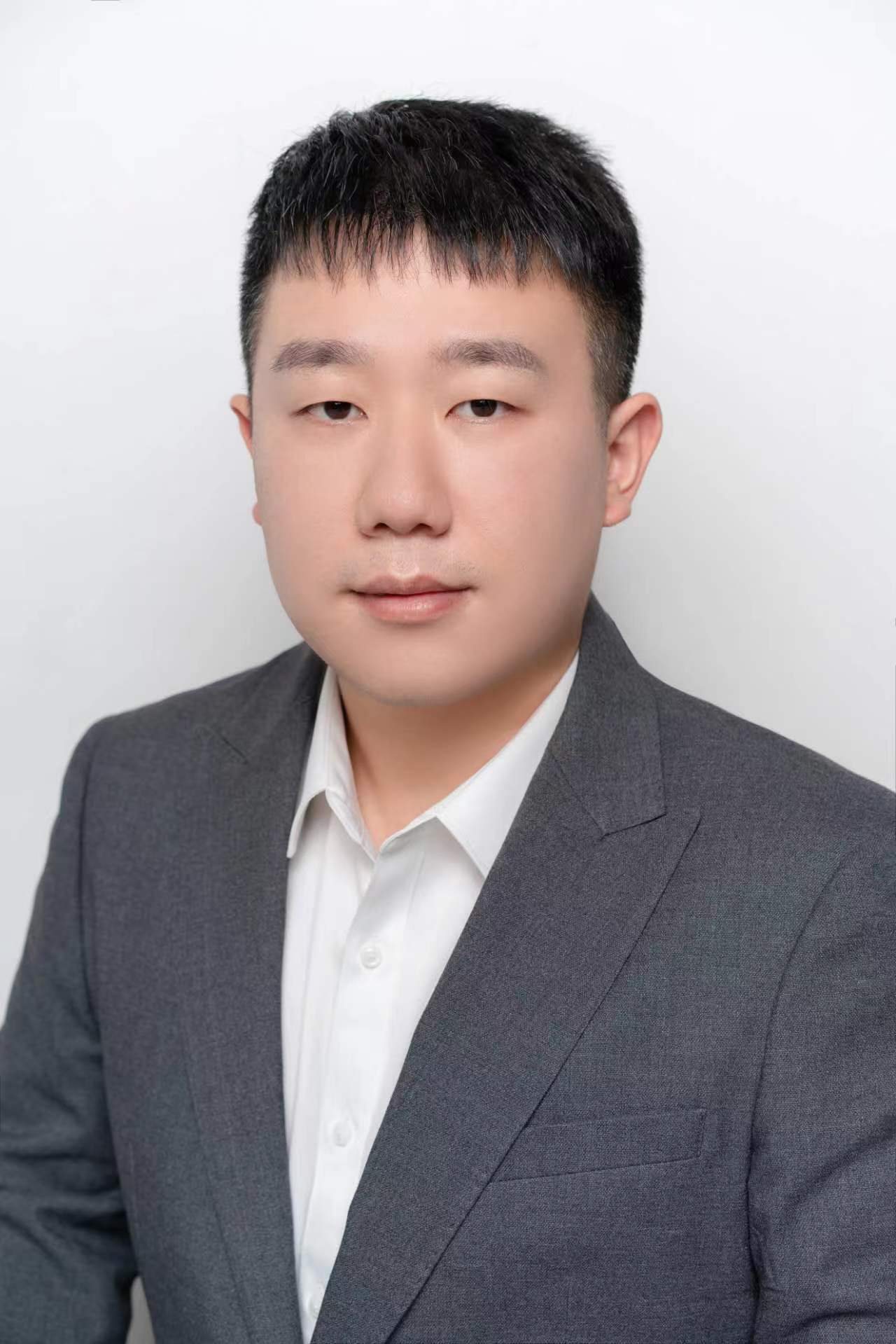}}
\noindent {\bf Tie Junbo} is currently a  research assistant in the College of Computer Science and Technology, National University of Defense Technology. He received the B.Eng. degree and the Ph.D. degree from National University of Defense Technology in 2013 and 2018, respectively. He current research interests include computer architecture, artificial intelligence, and neuromorphic computation.
}
\vspace{2\baselineskip}

\par\noindent 
\parbox[t]{\linewidth}{
\noindent\parpic{\includegraphics[height=1.5in,width=1in,clip,keepaspectratio]{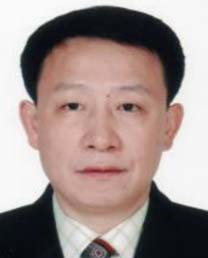}}
\noindent {\bf Xiaoping Hu} (Senior Member, IEEE) received the B.S. and M.S. degrees in automatic control systems and aircraft design from the College of Mechatronic Engineering and Automation, National University of Defense Technology, Changsha, China, in 1982 and 1985, respectively. He is currently a Professor with the National University of Defense Technology. His scientific interests include navigation, aircraft guidance, and control.
}
\vspace{2\baselineskip}

\par\noindent 
\parbox[t]{\linewidth}{
\noindent\parpic{\includegraphics[height=1.5in,width=1in,clip,keepaspectratio]{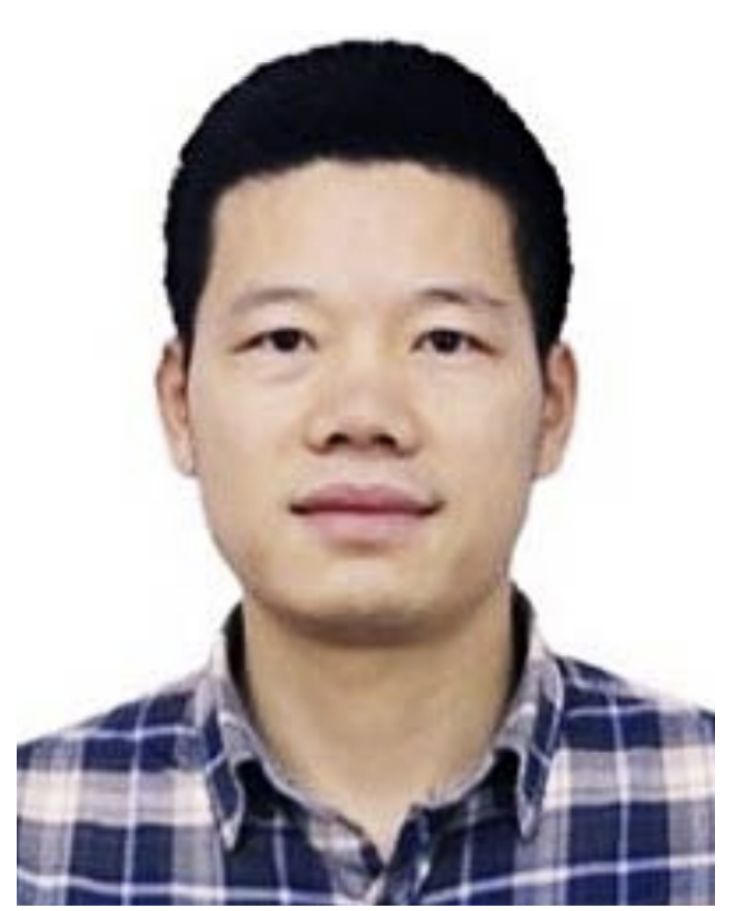}}
\noindent {\bf Xiaofeng He} received the B.S. and Ph.D. degrees from the School of Mechatronics and Automation, National University of Defense Technology, China, in 2001 and 2009, respectively.Since 2009, he has been an Associate Professor with the National University of Defense Technology. His current research interests include satellite navigation and deeply integrated navigation systems.
}
\vspace{2\baselineskip}

\par\noindent 
\parbox[t]{\linewidth}{
\noindent\parpic{\includegraphics[height=1.5in,width=1in,clip,keepaspectratio]{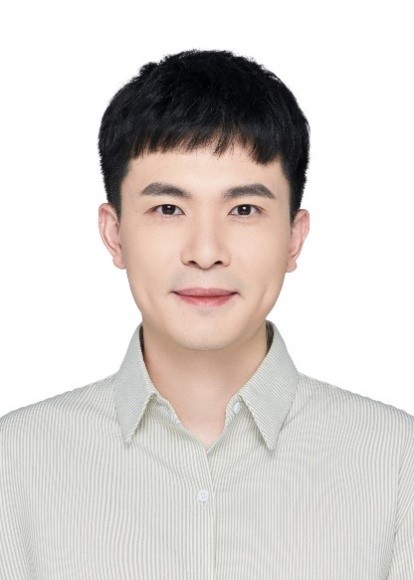}}
\noindent {\bf Yifei Shi} is an Associate Professor at the College of Intelligence Science and Technology, National University of Defense Technology (NUDT). He received his Ph.D. degree in computer science from NUDT in 2019. During 2017-2018, he was a visiting student research collaborator at Princeton University. His research interests mainly include computer vision, computer graphics, especially on object/scene analysis and manipulation by machine learning and geometric processing techniques. He has published 20+ papers in top-tier conferences and journals, including CVPR, ECCV, ICCV, SIGGRAPH Asia, IEEE Transactions on Pattern Analysis and Machine Intelligence, and ACM Transactions on Graphics.
}
\vspace{2\baselineskip}

\par\noindent 
\parbox[t]{\linewidth}{
\noindent\parpic{\includegraphics[height=1.5in,width=1in,clip,keepaspectratio]{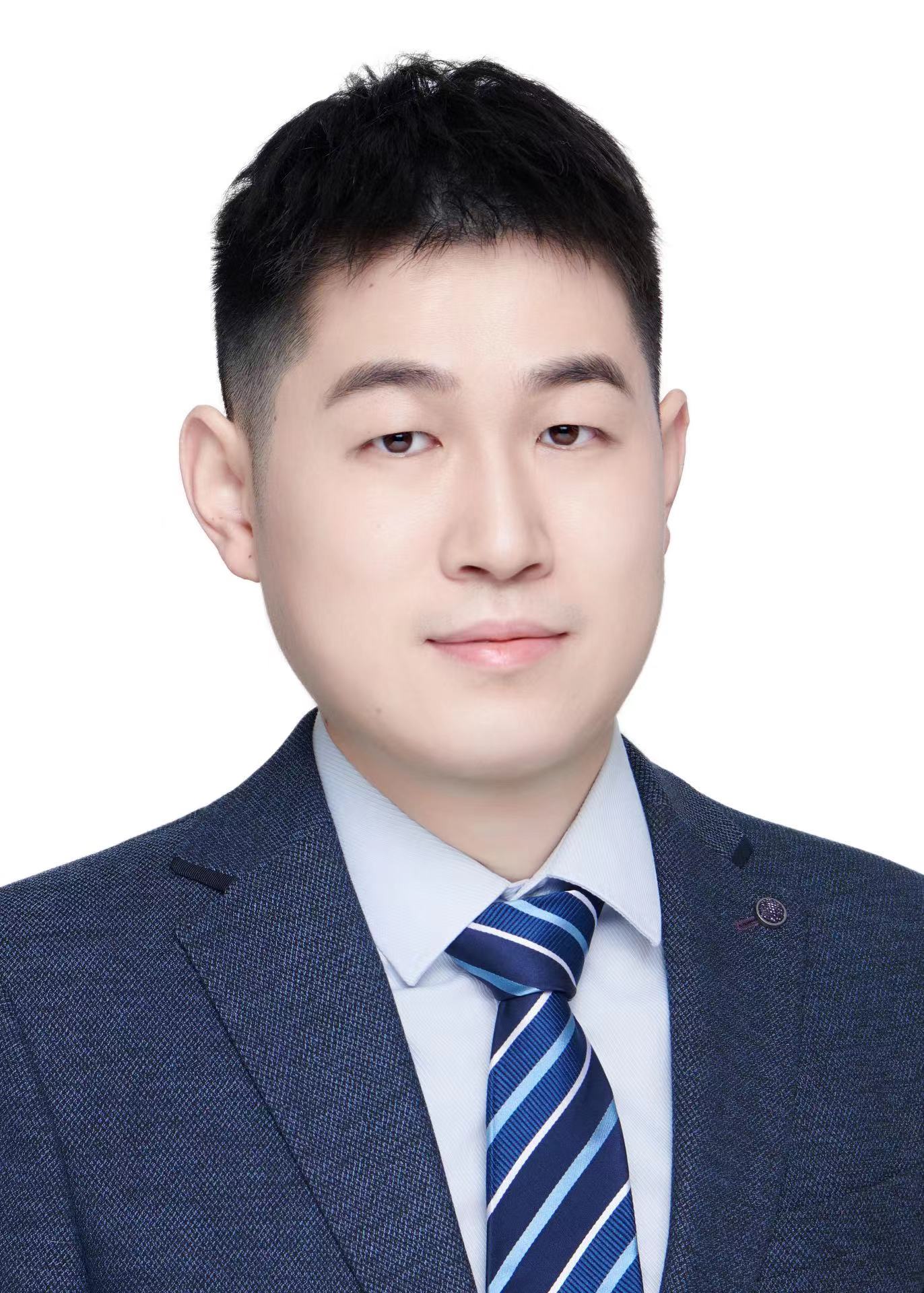}}
\noindent {\bf Changhao Chen} is an Assistant Professor at the College of Intelligence Science and Technology, National University of Defense Technology. Before that, he obtained his Ph.D. degree at University of Oxford (UK), M.Eng. degree at National University of Defense Technology (China), and B.Eng. degree at Tongji University (China). His research interest lies in robotics, computer vision and cyberphysical systems.
}
\vspace{2\baselineskip}

\end{document}